\documentclass[lettersize,journal]{IEEEtran}

 \usepackage{epsfig}
\usepackage{lineno}
\usepackage{booktabs}
\usepackage{makecell,multirow}
\usepackage{rotating}
\usepackage[table]{xcolor}
\usepackage{xcolor}
\usepackage[normalem]{ulem}
\usepackage[utf8]{inputenc}
\usepackage[T1]{fontenc}
\definecolor{mygreen}{rgb}{0.0, 0.6, 0.0}

\usepackage{amsmath,amsfonts}
\usepackage{algorithmic}
\usepackage{array}
\usepackage[caption=false,font=normalsize,labelfont=sf,textfont=sf]{subfig}
\usepackage{textcomp}
\usepackage{stfloats}
\usepackage{url}
\usepackage{verbatim}
\usepackage{graphicx}
\def\BibTeX{{\rm B\kern-.05em{\sc i\kern-.025em b}\kern-.08em
    T\kern-.1667em\lower.7ex\hbox{E}\kern-.125emX}}
\usepackage{balance}
\begin{document}
\title{FISformer: Replacing Self-Attention with a Fuzzy Inference System in Transformer Models for Time Series Forecasting}

\author{
    Bülent Haznedar and Levent Karacan
    \thanks{B. Haznedar and L. Karacan are with the Computer Engineering Department, Gaziantep University, Gaziantep, Türkiye. Corresponding author: Bülent Haznedar (e-mail: haznedar@gaziantep.edu.tr). 
    \\
    This work has been submitted to the IEEE for possible publication. Copyright may be transferred without notice, after which this version may no longer be accessible.}
}

\maketitle

\begin{abstract}
Transformers have achieved remarkable progress in time series forecasting, yet their reliance on deterministic dot-product attention limits their capacity to model uncertainty and nonlinear dependencies across multivariate temporal dimensions. To address this limitation, we propose FISFormer, a Fuzzy Inference System–Driven Transformer that replaces conventional attention with a FIS Interaction mechanism. In this framework, each query–key pair undergoes a fuzzy inference process for every feature dimension, where learnable membership functions and rule-based reasoning estimate token-wise relational strengths. These FIS-derived interaction weights capture uncertainty and provide interpretable, continuous mappings between tokens. A softmax operation is applied along the token axis to normalize these weights, which are then combined with the corresponding value features through element-wise multiplication to yield the final context-enhanced token representations. This design fuses the interpretability and uncertainty modeling of fuzzy logic with the representational power of Transformers. Extensive experiments on multiple benchmark datasets demonstrate that FISFormer achieves superior forecasting accuracy, noise robustness, and interpretability compared to state-of-the-art Transformer variants, establishing fuzzy inference as an effective alternative to conventional attention mechanisms.
\end{abstract}


\section{Introduction}

\IEEEPARstart{F}{orecasting} future trends from historical data in time-dependent systems is a well-established yet challenging task with significant implications across various domains, including energy management, traffic flow estimation, disease propagation, and finance. Over the past two decades, deep learning-based approaches, particularly recurrent neural networks (RNNs) such as Long Short-Term Memory (LSTM) networks and Gated Recurrent Units (GRUs) \cite{hochreiter1997long, cho2014properties, qin2017dual, salinas2020deepar, wen2017multi}, have been widely employed for time series forecasting. While these architectures effectively capture short-term dependencies, they often struggle with long-term forecasting when limited past information is available.

More recently, transformer-based models \cite{zhou2021informer, wu2021autoformer, zhou2022fedformer, zhang2023crossformer, liu2023itransformer} have demonstrated remarkable success in multivariate time series forecasting, primarily due to their ability to model long-range dependencies through self-attention mechanisms. However, the standard self-attention operation suffers from quadratic complexity concerning sequence length, posing a significant challenge for long-horizon predictions. To improve computational efficiency, several techniques have been proposed, including sparse attention \cite{li2019enhancing, child2019generating}, locality-sensitive hashing \cite{kitaev2019reformer}, low-rank projections \cite{wang2020linformer}, and probabilistic attention mechanisms \cite{zhou2021informer}. Beyond computational optimization, recent works have explored improvements beyond computational efficiency by enhancing multivariate dependency modeling. For instance, Crossformer \cite{zhang2023crossformer} introduces cross-dimensional dependencies, while iTransformer \cite{liu2023itransformer} replaces temporal tokens with variate tokens to improve multivariate forecasting performance.

Despite these advancements, traditional self-attention mechanisms rely on crisp similarity measures between query and key tokens, which may not fully capture the uncertainty and variability inherent in real-world time series data. Moreover, conventional attention assigns static importance scores based on pairwise token interactions, limiting its adaptability to dynamic dependencies.

To address these limitations, fuzzy-based extensions of Transformer architectures have recently been introduced. For instance, the Fuzzy Attention Transformer \cite{kong2024ficformer} and the Fuzzy Dynamic Transformer (FDformer) \cite{ren2025fdformer} incorporate fuzzy membership functions into the attention computation to improve robustness under noisy industrial and multivariate conditions. Engel \textit{et al.}~\cite{engel2024transformer} proposed a Transformer with a fuzzy attention mechanism in the context of weather time‐series forecasting, using a context vector to modulate attention via fuzzy measures. While effective for robustness and interpretability, the method retains the standard self‐attention similarity operator. Similarly, Chakraborty and Heintz \cite{chakraborty2025enhancing} proposed a Fuzzy Attention-Integrated Transformer (FANTF) that augments the standard dot-product attention with fuzzified similarity scores using differentiable fuzzification~\cite{zhan2023differential}. While these approaches demonstrate that fuzzy logic can enhance uncertainty modeling in Transformers, their integration remains largely superficial, as the fuzzy component is typically applied only as a weighting modifier or post-processing step within the attention layer. Consequently, they fail to exploit the full reasoning and rule-based potential of fuzzy inference systems (FIS).

In contrast to previous fuzzy-enhanced attention mechanisms that apply localized fuzzification to attention weights, we introduce FISformer, a fundamentally different architecture that replaces the standard similarity computation with a Fuzzy Inference System-Based Token Interaction module. Instead of merely modifying the attention scores, FISformer embeds a complete fuzzy inference process into the query–key interaction stage, grounded in the Sugeno-type fuzzy inference system \cite{takagi1985fuzzy}. In this formulation, learnable membership functions and interpretable fuzzy rules jointly infer the relational strength between tokens, producing continuous and differentiable similarity scores suitable for end-to-end optimization. This design enables nonlinear and uncertainty-aware similarity estimation while preserving the structural advantages of Transformer architectures. By grounding token interactions in rule-based reasoning rather than heuristic fuzzification, FISformer provides enhanced adaptability to dynamic dependencies, improved robustness under noisy conditions, and transparent interpretability through its rule activations.

The key contributions of this study are summarized as follows:
\begin{enumerate}

    \item FIS-based Token Interaction: We propose a novel interaction mechanism that embeds a complete Fuzzy Inference System (FIS) into the Transformer architecture. Instead of relying on conventional dot-product attention, token-to-token affinities are inferred through learnable fuzzy membership functions and rule-based reasoning, enabling nonlinear, uncertainty-aware, and interpretable similarity estimation.
    \item Learnable Membership and Rule Adaptation: The proposed model dynamically learns both the fuzzy membership parameters and the rule bases through end-to-end training, allowing adaptive mapping of query–key relationships under diverse temporal dynamics and feature scales.  

    \item Hybrid Transformer Design: FISformer retains the structural benefits of the Transformer while replacing its core similarity computation with fuzzy inference, combining the expressive representation power of deep attention-based architectures with the interpretability and robustness of fuzzy logic.  
    
    \item Comprehensive Experimental Validation: Extensive experiments on multiple multivariate time series benchmarks (e.g., ETT, ECL, Weather) demonstrate that FISformer consistently outperforms state-of-the-art Transformer variants in forecasting accuracy, noise robustness, and interpretability, while maintaining computational efficiency comparable to standard attention models.

\end{enumerate}

The proposed FISformer framework establishes a new paradigm in transformer-based time series forecasting by integrating fuzzy inference into the core of token interaction. Through its FIS-based Token Interaction mechanism, FISformer bridges the gap between interpretability, adaptability, and computational efficiency, providing a principled alternative to conventional self-attention. This integration of rule-based fuzzy reasoning with deep contextual representation not only enhances forecasting accuracy and robustness but also enables transparent and explainable temporal dependency modeling.

\section{Related Works}

\subsection{Time Series Forecasting}

Time series forecasting is a long-standing research problem with wide-ranging applications across various domains. Traditional approaches often relied on statistical models such as structural vector autoregression and multivariate dynamic linear models \cite{lutkepohl2006structural, prado2006multivariate}. For long-term forecasting, early work such as \cite{sorjamaa2007methodology} introduced novel input selection strategies to improve least squares support vector machines.

With the advent of deep learning, data-driven methods have become dominant in time series forecasting. A prominent line of research has focused on recurrent neural networks (RNNs) \cite{yu2017long, wen2017multi, rangapuram2018deep, maddix2018deep}. DeepAR \cite{salinas2020deepar} leveraged autoregressive RNNs to learn a global probabilistic forecasting model from a collection of related time series. LSTNet \cite{lai2018modeling} combined convolutional and recurrent components to capture both short- and long-term dependencies in multivariate time series. Similarly, attention-enhanced RNN models have been explored to improve interpretability and performance \cite{qin2017dual, song2018attend, shih2019temporal}.

In addition to RNN-based methods, filtering techniques have been integrated into deep learning frameworks for multivariate forecasting \cite{de2020normalizing, kurle2020deep}. Temporal convolutional networks (TCNs) have also gained attention for their ability to model long-range dependencies efficiently \cite{borovykh2017conditional, bai2018empirical, sen2019think}.

Recently, transformer-based architectures have become increasingly popular in time series forecasting. Much of the research in this area focuses on modifying the attention mechanism or redesigning the overall architecture \cite{liu2023itransformer}. Informer \cite{zhou2021informer} introduces the ProbSparse attention mechanism, which significantly reduces computational complexity for long sequences while preserving effective dependency modeling. Autoformer \cite{wu2021autoformer} presents an auto-correlation-based architecture that incorporates a seasonal-trend decomposition mechanism. Similarly, FEDformer \cite{zhou2022fedformer} enhances forecasting performance by integrating frequency-domain decomposition with the transformer framework.

Zeng et al. \cite{zeng2023transformers} critically evaluate transformer models for time series forecasting and highlight their limitations, showing that standard designs may underperform compared to simpler models. Liu et al. \cite{liu2022non} address non-stationarity in time series by proposing Non-stationary Transformers, which adapt to dynamic statistical properties and improve long-term accuracy. Nie et al. \cite{nie2022time} introduce a compact transformer model that treats time series as token sequences, enabling efficient long-horizon forecasting.

Crossformer \cite{zhang2023crossformer} enhances forecasting for multivariate time series by explicitly modeling cross-dimensional dependencies through a novel attention mechanism and architectural design. Most recently, Liu et al. \cite{liu2023itransformer} propose iTransformer, which inverts the standard transformer structure while preserving its core components, yielding improvements in efficiency and long-term forecasting performance.

\subsection{Fuzzy Systems with Deep Learning for Time Series}

Several studies have demonstrated the effectiveness of FIS in time series forecasting. For instance, AutoMFIS \cite{coutinho2016automfis} and its extensions \cite{carvalho2023automatic} applied fuzzy reasoning in conjunction with classical models for multivariate forecasting tasks. More recently, FIS has also been integrated with recurrent architectures. Li et al. (2020) \cite{li2020t2f} proposed T2F-LSTM, combining type-2 fuzzy logic with LSTM networks to better capture uncertainty and non-stationarity in traffic volume prediction. Tang et al. (2021) \cite{tang2021building} introduced a trend granulation-based fuzzy-LSTM model that transforms time series into fuzzy information granules to enhance long-term forecasting. Similarly, Wang et al. (2023) \cite{wang2023fuzzy} developed a fuzzy inference-based LSTM that improves prediction performance under noisy and uncertain conditions through rule-based reasoning.

In recent years, fuzzy-augmented Transformers have emerged for time-series forecasting. Models such as Fuzzy Attention Transformer \cite{kong2024ficformer}, FDformer \cite{ren2025fdformer}, and FANTF \cite{chakraborty2025enhancing} introduce fuzzy membership functions \cite{zhan2023differential} into attention computation to improve robustness under noisy or uncertain conditions. Engel \textit{et al.}~\cite{engel2024transformer} proposed a Transformer with a fuzzy attention mechanism in the context of weather time‐series forecasting, using a context vector to modulate attention via fuzzy measures. While effective, these approaches typically modify attention scores or feature embeddings rather than replacing the core token interaction mechanism. Consequently, the potential of full rule-based fuzzy inference to model per-token, per-dimension interactions remains largely unexplored.

Despite the extensive development of neural and Transformer-based models for time series forecasting, Fuzzy Inference Systems (FIS) have not, to the best of our knowledge, been directly incorporated into modern Transformer architectures. FISs provide a mathematically grounded framework for reasoning under uncertainty and deliver human-like interpretability through linguistic rule bases. These properties make them particularly effective for modeling noisy, ambiguous, and nonlinear temporal dependencies, ubiquitous in real-world multivariate time series. In this work, we address this gap by proposing FISformer, a novel Transformer architecture driven by a Sugeno-type Fuzzy Inference System \cite{takagi1985fuzzy}. Rather than applying shallow fuzzification to attention weights, FISformer integrates a FIS-based Token Interaction mechanism that replaces the conventional dot-product similarity between queries and keys with rule-based fuzzy inference. This design unifies the interpretability and robustness of fuzzy logic with the expressive representation power of Transformers, providing a principled and uncertainty-aware alternative to conventional self-attention for time series forecasting.

\section{Method}

Given a multivariate time series input \( X=\{x_1,...,x_T\} \in \mathbb{R}^{T \times N } \), where \( N \) denotes the number of variates, \( T \) represents the number of past time steps, the objective is to forecast the corresponding target output \( Y=\{y_{T+1},...,y_{T+P}\} \in \mathbb{R}^{P \times N} \), where \( P \) is the number of future time steps to be predicted. This forecasting task is addressed by learning a mapping function:

\begin{equation}
Y = \mathcal{F}(X)
\end{equation}

where \( \mathcal{F} \) represents the predictive model trained to capture temporal and cross-variable dependencies within the input data.

\subsection{Network Architecture}

\begin{figure*}[ht!]
    \centering
    
    \includegraphics[width=1.0\textwidth]{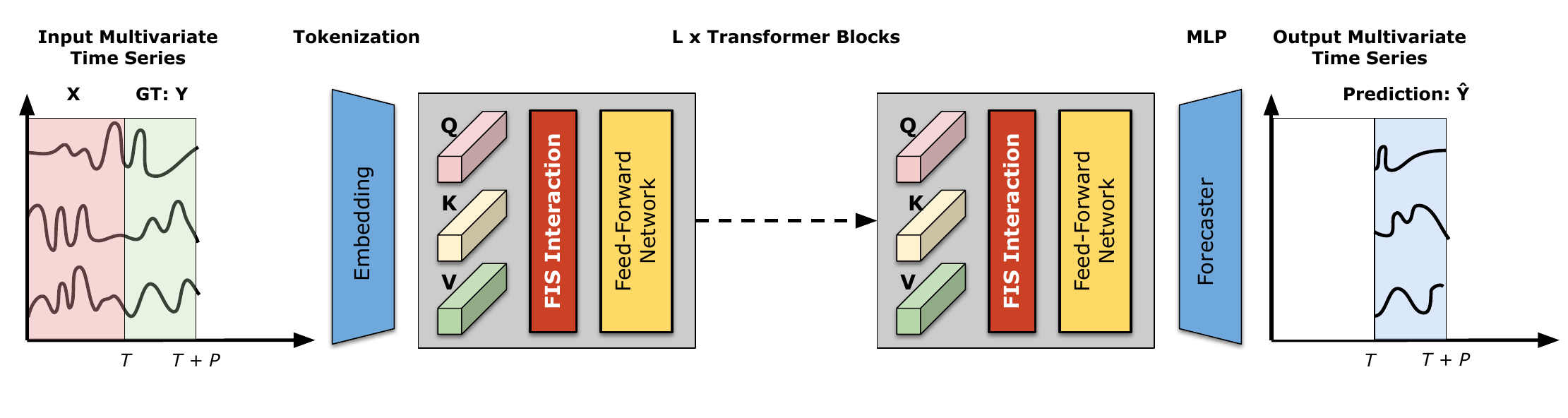} 
    \caption{Overall architecture of the proposed model, built upon a Transformer encoder. It comprises an embedding layer to convert multivariate input sequences into dense representations, a Transformer block to capture temporal and feature-wise dependencies via the proposed FIS-based Token Interaction module, and a projection module to map encoded features to the prediction output.}
    \label{fig:overallarch}
\end{figure*}
The overall architecture of the proposed model is illustrated in Figure~\ref{fig:overallarch}.  The model is built upon a Transformer encoder framework, which comprises three main components: an embedding layer, $L$ Transformer blocks, and a projection module. The embedding layer is responsible for mapping the input time series into a suitable representation space, the Transformer block captures complex temporal and cross-variable dependencies through self-attention mechanisms, and the projection module generates the final output by mapping the encoded features to the target prediction space.

There are several approaches to embedding raw time series data into the representation space. In the first approach adopted by models such as Informer~\cite{zhou2021informer}, Autoformer~\cite{wu2021autoformer}, Crossformer~\cite{zhang2023crossformer}, and FEDformer~\cite{zhou2022fedformer}, a token is constructed for each time step, where each variable (or feature) within the multivariate time series is treated as a separate token. This method allows the model to capture temporal dynamics across individual variables. In the second approach, as in iTransformer~\cite{liu2023itransformer}, each variable is assigned a distinct token across all time steps, which helps to mitigate inter-variable interference by decoupling the representations of different variates. This strategy aims to enhance the model's ability to learn variable-specific patterns without introducing noise from other variables.

In this study, we adopt the second embedding strategy, in which each variate in the multivariate time series is treated as an independent token. However, our proposed approach is not inherently constrained by this tokenization scheme; it also remains compatible with alternative strategies such as time-step-based tokenization that are employed in prior works. Given an input multivariate time series \( X \in \mathbb{R}^{T \times N} \), where \( T \) denotes the number of time steps and \( N \) represents the number of  variates, the input is embedded into a high-dimensional latent space of dimension \( D \) using the following procedure:
\begin{equation}
   \begin{split}
    H^0 &= \text{Embedding}(X), \\
    H^{l+1} &= \text{TrmBlock}(H^l), \quad l = 0, \ldots, L-1, \\
    \hat{Y} &= \text{Projection}(H^L),
    \end{split} 
\end{equation}
where \( H^l \in \mathbb{R}^{N \times D} \) denotes the intermediate hidden representation at the \( l \)-th layer, \( N \) is the number of tokens (equal to \( d \) under variate-wise tokenization), and \( L \) is the total number of stacked Transformer layers. Each \texttt{TrmBlock} consists of a standard Transformer encoder block, which includes proposed FIS-based token interaction layer, layer normalization, and a feed-forward network with residual connections. The embedding layer projects each variate into a shared latent space, enabling the model to capture complex inter-variable relationships. Finally, the \texttt{Projection} module, typically implemented as a linear transformation, maps the final layer output \( H^L \) into the prediction space, producing the forecast \( \hat{Y} \in \mathbb{R}^{P \times N} \), where \( P \) is the number of future time steps to predict.

\subsection{FIS-based Token Interaction Layer}

\begin{figure*}[ht]
    \centering
    \includegraphics[width=1.0\textwidth]{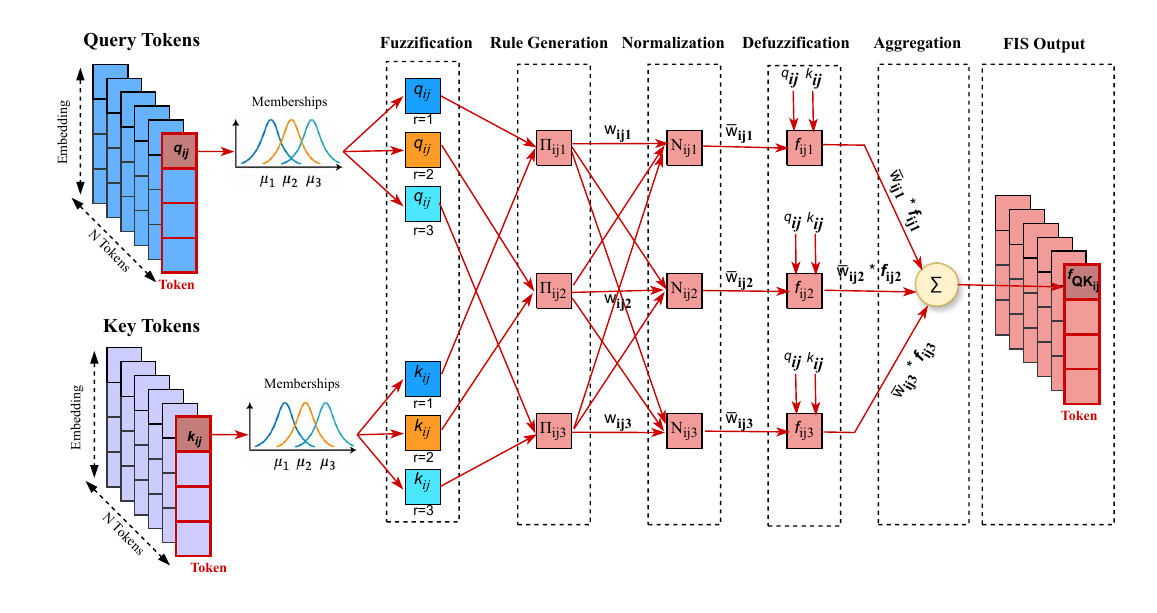} %
    \caption{FIS-based Token Interaction. Query and key tokens are fuzzified through learnable Gaussian membership functions and combined via rule-based fuzzy inference. The resulting fuzzy rules are defuzzified following a first-order Sugeno process to produce the Fuzzy Interaction Map.}

    \label{fig:fis}
\end{figure*}

\begin{figure}[ht]
    \centering
    \label{fig:fuzzyattention}
    \includegraphics[width=0.5\textwidth]{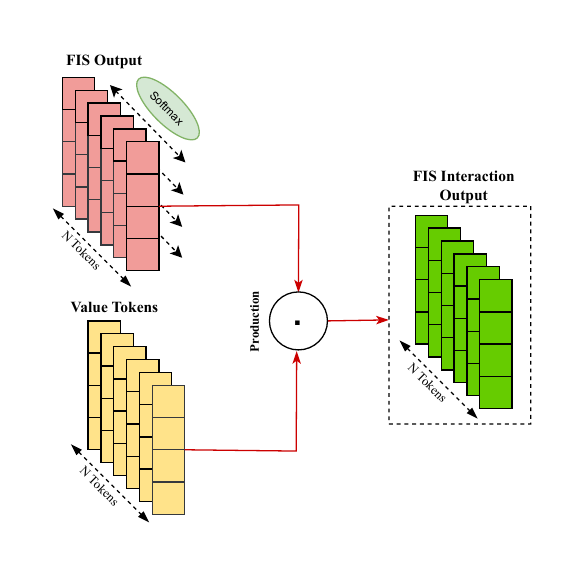} %
    \caption{FIS Interaction Output. The Fuzzy Interaction Map encodes inferred relational strengths across all token dimensions. A softmax operation is applied along the token axis to obtain normalized interaction weights, which are then combined with value representations via element-wise multiplication to produce context-enhanced token embeddings.}
    \label{fig:fattn}
\end{figure}

\textit{Preliminaries of Self-Attention.} The original self-attention mechanism was proposed to enable global interactions across temporal or spatial dimensions by computing pairwise relationships among tokens. Formally, given an input embedding matrix \( H \in \mathbb{R}^{T \times D} \), it is linearly projected into the representations of queries \( \mathbf{Q} \in \mathbb{R}^{T \times d} \), keys \( \mathbf{K} \in \mathbb{R}^{T \times d} \), and values \( \mathbf{V} \in \mathbb{R}^{T \times d} \), each with dimensionality \( d \). The attention mechanism computes interactions between query and key tokens through scaled dot-product attention, defined as
\begin{equation}
A_{i,j} = \left( \frac{QK^\top}{\sqrt{d}} \right)_{i,j} \propto q_i^\top k_j,
\end{equation}
where \( A \in \mathbb{R}^{T \times T} \) denotes the attention matrix capturing the relevance between each pair of tokens. This attention matrix is then used to aggregate contextual information by weighting the value tokens, resulting in the output representation
\begin{equation}
H' = AV.
\end{equation}

\textit{FIS-based Token Interaction.}
While self-attention enables efficient global interaction by capturing long-range dependencies, it relies on deterministic dot-product similarity, which lacks the capability to represent uncertainty and nonlinear relations inherent in real-world multivariate time series. 
To overcome these limitations, we introduce a novel \textit{Fuzzy Inference System (FIS)--based Interaction} mechanism that replaces the standard similarity computation with a rule-driven fuzzy reasoning process. 
The proposed module is inspired by the structure of a first-order Sugeno fuzzy inference system and operates through four key stages that collectively emulate the fuzzy reasoning pipeline (see Figure~\ref{fig:fis}): (i)~\textit{Fuzzification:} the input query and key embeddings are mapped to fuzzy membership degrees using learnable Gaussian membership functions, representing the degree of activation for each fuzzy set; (ii) \textit{Rule Generation:} Fuzzy rules are formulated over the combinations of fuzzified query and key memberships, where each rule represents an interpretable interaction between tokens. The activation of all rules is then normalized across the rule dimension to form a stable weighting distribution, ensuring balanced contribution of each rule in the subsequent defuzzification stage; (iii)~\textit{Defuzzification and Aggregation:} each fuzzy rule produces a numerical consequent based on the query--key inputs, consistent with a first-order Sugeno fuzzy inference model. (iv)~\textit{FIS Interaction:} the resulting \textit{Fuzzy Interaction Map} $\mathbf{F}_{QK}$ contains the inferred relational strengths across all token dimensions. 
For each dimension, a softmax operation is applied along the token axis to obtain normalized interaction weights, which are then combined with the corresponding value representations through element-wise multiplication to generate the final context-enhanced token embeddings. The resulting \textit{Fuzzy Interaction Map} $\mathbf{F}_{QK}$ is then normalized via a softmax operation and multiplied by the value representations to generate context-enriched token embeddings.

\subsubsection*{(i) Fuzzification}
Given a query matrix \( Q \in \mathbb{R}^{T \times d} \) and a key matrix \( K \in \mathbb{R}^{T \times d} \), each feature value is fuzzified using membership functions (MFs) to obtain fuzzy degrees of membership. In this study, we utilize three Gaussian membership functions per feature, each parameterized by learnable centers and standard deviations. Specifically, for each temporal index \( i \in \{1, 2, \ldots, T\} \), feature dimension \( j \in \{1, 2, \ldots, d\} \), and membership function index \( r \in \{1, 2, 3\} \), the fuzzy membership degrees for the query and key are computed as:
\begin{equation}
\begin{split}
\mu_q(q_{i,j}; r) &= \exp\left( -\frac{(q_{i,j} - w^Q_{i,j,r})^2}{2 (\sigma^Q_{i,j,r})^2} \right), \\
\mu_k(k_{i,j}; r) &= \exp\left( -\frac{(k_{i,j} - w^K_{i,j,r})^2}{2 (\sigma^K_{i,j,r})^2} \right),
\end{split}
\end{equation}
where \( w^Q_{i,j,r} \) and \( \sigma^Q_{i,j,r} \) denote the center and standard deviations of the \( r \)-th Gaussian MF for the \((i,j)\)-th element of the query, and \( w^K_{i,j,r} \), \( \sigma^K_{i,j,r} \) are the corresponding parameters for the key. These MF parameters are trainable and optimized jointly with the overall model. As a result, both query and key matrices are transformed into fuzzified representations of shape \( T \times d \times 3 \), effectively capturing soft and interpretable fuzzy memberships for each input feature.
\\
\subsubsection*{(ii) Rule Generation}
For the rule generation stage, we adopt a product-based t-norm to compute the activation strength of each fuzzy rule. In this framework, each rule node represents the interaction between the fuzzified query and key membership values generated during the fuzzification process. The degree to which a rule is activated—referred to as the firing strength—is determined by the product of the corresponding membership degrees. Specifically, for each time step \( i \in \{1, \ldots, T\} \), feature dimension \( j \in \{1, \ldots, d\} \), and membership function index \( r \in \{1, 2, 3\} \), the firing strength \( \pi_{i,j,r} \) is computed as:
\begin{equation}
\pi_{i,j,r} = \mu_q(q_{i,j}; r) \cdot \mu_k(k_{i,j}; r).
\end{equation}

where \( \mu_q(q_{i,j}; r) \) and \( \mu_k(k_{i,j}; r) \) denote the Gaussian membership values of the query and key, respectively. This operation effectively captures the joint activation level of the corresponding fuzzy rule for the given input features.

This corresponds to a first-order Sugeno fuzzy rule of the form
\[
\begin{aligned}
\text{IF } & q_{i,j} \text{ is } A_r \text{ AND } k_{i,j} \text{ is } B_r, \\
\text{THEN } & c_{i,j,r} = w_{r,q} \, q_{i,j} + w_{r,k} \, k_{i,j} + b_r.
\end{aligned}
\]
where \(A_r\) and \(B_r\) denote the fuzzy sets induced by the corresponding membership functions for the query and key, respectively.
\\
\subsubsection*{(iii) Defuzzification and Aggregation}

In the defuzzification stage, the activations of all fuzzy rules for each query--key pair $(i,j)$ are first normalized to form a valid weighting distribution. 
Let $\pi_{i,j,r}$ denote the firing strength of the $r$-th rule, and let $R$ be the total number of rules. 
The normalized activation $\tilde{\pi}_{i,j,r}$ is computed as
\begin{equation}
\tilde{\pi}_{i,j,r} = 
\frac{\pi_{i,j,r}}{\displaystyle \sum_{s=1}^{R} \pi_{i,j,s} + \varepsilon},
\label{eq:norm_firing_scalar}
\end{equation}
where $\varepsilon > 0$ is a small constant added for numerical stability.

Unlike the zero-order formulation, where each rule output is a fixed learnable constant, here each rule produces an \textit{input-dependent consequent} based on the scalar query and key inputs. 
For the $(i,j)$-th interaction, the augmented input vector is defined as
\begin{equation}
z_{i,j} = [\, q_{i,j}, \, k_{i,j}, \, 1 \,],
\label{eq:z_scalar}
\end{equation}
and each rule $r$ is parameterized by a learnable coefficient vector 
$\mathbf{w}_r = [\, w_{r,q}, \, w_{r,k}, \, b_r \,]^{\top}$.
The rule consequent is then computed as
\begin{equation}
c_{i,j,r} = w_{r,q} \, q_{i,j} + w_{r,k} \, k_{i,j} + b_r.
\label{eq:rule_consequent_scalar}
\end{equation}

Finally, the defuzzified fuzzy interaction score between the $i$-th query and $j$-th key is obtained through the weighted aggregation of all rule consequents:
\begin{equation}
f_{QK}(i,j) = \sum_{r=1}^{R} \tilde{\pi}_{i,j,r} \, c_{i,j,r},
\label{eq:defuzz_final_scalar}
\end{equation}
where $f_{QK}(i,j)$ represents the corresponding element of the \textit{Fuzzy Interaction Map}, serving as the output of the FIS.
\\
\subsubsection*{(iv) FIS Interaction}

Following the computation of the fuzzy inference outputs 
\( f_{QK} \in \mathbb{R}^{T \times d} \),
the model performs token-wise normalization and interaction with the value matrix. 
For each feature dimension, a softmax operation is applied along the token axis, to produce dimension-specific FIS interaction weights.
Formally, the normalized fuzzy weights are computed as
\begin{equation}
A_{:,j} = 
\text{softmax}\!\big(f_{QK}(:,j)\big),
\label{eq:fis_softmax}
\end{equation}
where \(A_{:,j} \in \mathbb{R}^{T}\) denotes the normalized interaction weights for the 
$j$-th feature dimension, and the softmax operation is applied along the token axis. 

The resulting weights determine the relative contribution of each token for every feature dimension. 
To obtain the final context-enhanced representation, the \textit{FIS interaction weights} are combined 
with the corresponding value features through element-wise multiplication:
\begin{equation}
O_{i,j} = A_{i,j} \cdot V_{i,j}, 
\quad \text{for } i = 1,2,\ldots,T, \;
j = 1,2,\ldots,d,
\label{eq:fis_elementwise}
\end{equation}
where \(V \in \mathbb{R}^{T \times d}\) denotes the value matrix 
and \(O \in \mathbb{R}^{T \times d}\) represents the final output of the 
\textit{FIS Interaction} layer.

This formulation applies a fuzzy inference process at each feature dimension,
yielding interpretable and uncertainty-aware \textit{FIS interaction weights} across tokens.
Unlike conventional self-attention, which aggregates global correlations through dot-product similarity, 
the FIS Interaction mechanism derives token contributions directly from fuzzy rule activations, 
offering a continuous, differentiable, and interpretable mapping between query--key pairs and their 
resulting contextual representations.

\section{Experiments}
We conduct a comprehensive evaluation of our proposed model across a variety of time series forecasting tasks to validate the generality and robustness of the framework. We also evaluate the impact of integrating the proposed FIS interaction module into a sample Transformer-based model, highlighting the benefits of the proposed mechanism.

\subsection{Dataset Descriptions}

\begin{table*}[h!]
    \centering
    \caption{Detailed dataset descriptions including input dimensions, number of samples, prediction sequence length, and time frequency.}
    \label{tab:dataset}
    \scalebox{1}{ 
    \setlength{\tabcolsep}{10pt}
    \begin{tabular}{c | c | c | c | c | c }
        \toprule
        Dataset & Dim & Prediction Length & Dataset Size & Frequency & Information \\
        \midrule
        \makecell{ETTh1, ETTh2} & 7 & \{96, 192, 336, 720\} & (8545, 2881, 2881) & Hourly & Electricity \\
        \makecell{ETTm1, ETTm2} & 7 & \{96, 192, 336, 720\} & (34465, 11521, 11521) & 15min & Electricity \\
        \makecell{Exchange} & 8 & \{96, 192, 336, 720\} & (5120, 665, 1422) & Daily & Economy \\
        \makecell{Weather} & 21 & \{96, 192, 336, 720\} & (36792, 5271, 10540) & 10min & Weather \\
        \makecell{ECL} & 321 & \{96, 192, 336, 720\} & (18317, 2633, 5261) & Hourly & Electricity \\
        \makecell{Traffic} & 862 & \{96, 192, 336, 720\} & (12185, 1757, 3509) & Hourly & Transportation \\
        \makecell{Solar-Energy} & 137 & \{96, 192, 336, 720\} & (36601, 5161, 10417) & 10min & Energy \\
        \makecell{PEMS03} & 358 & \{12, 24, 48, 96\} & (15617, 5135, 5135) & 5min & Transportation \\
        \makecell{PEMS04} & 307 & \{12, 24, 48, 96\} & (10172, 3375, 3375) & 5min & Transportation \\
        \makecell{PEMS07} & 883 & \{12, 24, 48, 96\} & (16911, 5622, 5622) & 5min & Transportation \\
        \makecell{PEMS08} & 170 & \{12, 24, 48, 96\} & (10690, 3548, 3548) & 5min & Transportation \\
        \bottomrule
    \end{tabular}
    }
\end{table*}

The experimental studies are conducted on seven real-world datasets to evaluate the performance of the our proposed model. These datasets span a range of domains, including energy, finance, meteorology, and traffic. The ETT dataset~\cite{zhou2021informer} contains 7 features related to the temperature of electrical transformers, collected from July 2016 to July 2018, and is divided into four subsets: ETTh1 and ETTh2 (recorded hourly) and ETTm1 and ETTm2 (recorded every 15 minutes). The Exchange dataset~\cite{wu2021autoformer} includes daily exchange rate data from eight countries spanning the years 1990 to 2016. The Weather dataset~\cite{wu2021autoformer} comprises 21 meteorological variables recorded every 10 minutes in 2020 by the Weather Station at the Max Planck Institute for Biogeochemistry. The ECL dataset~\cite{wu2021autoformer} consists of hourly electricity consumption data from 321 clients. The Traffic dataset~\cite{wu2021autoformer} provides hourly road occupancy rates measured by 862 sensors across freeways in the San Francisco Bay Area between January 2015 and December 2016. The Solar-Energy dataset~\cite{lai2018modeling} captures solar power generation from 137 photovoltaic plants in 2006, sampled at 10-minute intervals. Lastly, the PEMS dataset contains public traffic network data from California, collected at 5-minute intervals; we use the four public subsets (PEMS03, PEMS04, PEMS07, and PEMS08) previously adopted in SCINet~\cite{liu2022non}.

We adopt the same data preprocessing and train-validation-test splitting strategy as proposed in TimesNet (Wu et al., 2023), where the datasets are partitioned strictly in chronological order to prevent any potential data leakage. For forecasting tasks, the lookback window length is fixed at 96 time steps across the ETT, Weather, ECL, Solar-Energy, PEMS, and Traffic datasets. The prediction horizon varies among four settings: {96, 192, 336, 720}. In the case of the PEMS dataset, following the setup used in SCINet—the prior state-of-the-art model—the prediction lengths are set to {12, 24, 48, 96}. A comprehensive summary of the datasets is provided in Table~\ref{tab:dataset}. In here, \textit{Dim} refers to the number of variables (or features) in each dataset. \textit{Dataset Size} indicates the total number of time points, presented as a tuple corresponding to the training, validation, and test splits, respectively. \textit{Prediction Length} denotes the number of future time points to be forecasted with four 4 prediction settings applied to each dataset. \textit{Frequency} represents the sampling interval at which time points are recorded. These specifications provide a comprehensive overview of the datasets and ensure consistency across experimental settings.
\subsection{Implementation Details}
All experiments are implemented using the PyTorch framework and executed on a single NVIDIA GeForce RTX 2080 Ti GPU with 12GB of memory. Model optimization is carried out using the Adam optimizer, with the initial learning rate selected from the set \( \{10^{-4}, 3 \times 10^{-4}, 5 \times 10^{-4}, 10^{-3} \} \), and the loss function employed is the standard L2 loss. A fixed batch size of 32 is used across all experiments, and training is performed for 10 epochs. The number of inverted Transformer blocks \( L \) in the proposed architecture is chosen from \( \{2, 3, 4\} \), and the dimensionality of the time series representations \( D \) is selected from \( \{256, 512\} \). For fair comparison, all baseline models are reproduced based on the official implementation and configuration guidelines provided in the TimesNet repository, which adheres to each model’s original paper or released codebase.

\subsection{Forecasting Results}

\begin{table*}[h!]
    \centering
    \caption{Average forecasting performance in terms of MSE and MAE across different prediction lengths. The results are reported for sequence lengths \{12, 24, 48, 96\} on the PEMS dataset, and \{96, 192, 336, 720\} on the other datasets. The best results are highlighted in \textcolor{red}{red}, and the second-best results are underlined \textcolor{blue}{blue}.}

    \label{tab:results}
   
    \scalebox{1}{ 
    \begin{tabular}{c | c c | c c | c c | c c | c c | c c}
        \toprule
        \multirow{2}{*}{Models} &  \multicolumn{2}{c}{FISformer} & \multicolumn{2}{c}{ITransformer} & \multicolumn{2}{c}{Crossformer} & \multicolumn{2}{c}{Fedformer} & \multicolumn{2}{c}{Autoformer} & \multicolumn{2}{c}{Informer} \\
        & \multicolumn{2}{c}{(Ours)} & \multicolumn{2}{c}{(2024)} & \multicolumn{2}{c}{(2023)} & \multicolumn{2}{c}{(2022)} & \multicolumn{2}{c}{(2021)} & \multicolumn{2}{c}{(2021)} \\
        \cmidrule(lr){2-3} \cmidrule(lr){4-5} \cmidrule(lr){6-7} \cmidrule(lr){8-9} \cmidrule(lr){10-11} \cmidrule(lr){12-13}
        Metric & MSE & MAE & MSE & MAE & MSE & MAE & MSE & MAE & MSE & MAE & MSE & MAE \\
        \midrule
        \makecell{ECL}       & \textbf{\textcolor{red}{0.172}} & \textbf{\textcolor{red}{0.262}} & \underline{\textcolor{blue}{0.178}} & \underline{\textcolor{blue}{0.270}} & 0.244 & 0.334 & 0.214 & 0.327 & 0.227 & 0.338 & 0.392 & 0.448 \\
        \midrule
        \makecell{ETT (Avg)} & \textbf{\textcolor{red}{0.375}} & \textbf{\textcolor{red}{0.395}} & \underline{\textcolor{blue}{0.383}} & \underline{\textcolor{blue}{0.399}} & 0.685 & 0.578 & 0.407 & 0.428 & 0.465 & 0.459 & 1.342 & 0.859 \\
        \midrule
        \makecell{Exchange}  & \textbf{\textcolor{red}{0.356}} & \textbf{\textcolor{red}{0.402}} & \underline{\textcolor{blue}{0.360}} & \underline{\textcolor{blue}{0.403}} & 0.940 & 0.707 & 0.519 & 0.429 & 0.613 & 0.539 & - & - \\
        \midrule
        \makecell{Traffic}   & \textbf{\textcolor{red}{0.417}} & \textbf{\textcolor{red}{0.279}} & \underline{\textcolor{blue}{0.428}} & \underline{\textcolor{blue}{0.282}} & 0.550 & 0.304 & 0.610 & 0.376 & 0.628 & 0.379 & - & - \\
        \midrule
        \makecell{Weather}   & \textbf{\textcolor{red}{0.256}} & \textbf{\textcolor{red}{0.277}} & \underline{\textcolor{blue}{0.258}} & \underline{\textcolor{blue}{0.278}} & 0.259 & 0.315 & 0.309 & 0.360 & 0.338 & 0.382 & 0.574 & 0.552 \\
        \midrule
        \makecell{Solar-Energy} & \textbf{\textcolor{red}{0.232}} & \textbf{\textcolor{red}{0.261}} & \underline{\textcolor{blue}{0.233}} & \underline{\textcolor{blue}{0.262}} & 0.641 & 0.639 & 0.291 & 0.381 & 0.885 & 0.711 & - & - \\
        \midrule
        \makecell{PEMS (Avg)} & \underline{\textcolor{blue}{0.122}} & \underline{\textcolor{blue}{0.219}} & \textbf{\textcolor{red}{0.119}} & \textbf{\textcolor{red}{0.218}} & 0.220 & 0.304 & 0.224 & 0.327 & 0.614 & 0.575 & - & - \\
        \bottomrule
    \end{tabular}
    }
\end{table*}

\begin{table*}[ht!]
     \label{tab:results_detailed}
    \centering
    \caption{Detailed forecasting comparison in terms of MSE and MAE across different prediction lengths. Results are reported for sequence lengths  \{96, 192, 336, 720\} on the ETT, ECL, Exchange, Traffic, Weather, and Solar-Energy datasets. The best results are highlighted in \textcolor{red}{red}, and the second-best results are \underline{\textcolor{blue}{underlined blue}}.}

    \label{tab:all-results}
    \scalebox{1}{ 
    \setlength{\tabcolsep}{10pt}
    \begin{tabular}{p{0.2cm} | c | c c | c c | c c | c c | c c | c c}
        \toprule
        \multicolumn{2}{c|}{\multirow{2}{*}{Models}} &  \multicolumn{2}{c}{FISformer} & \multicolumn{2}{c}{ITransformer} & \multicolumn{2}{c}{Crossformer} & \multicolumn{2}{c}{Fedformer} & \multicolumn{2}{c}{Autoformer} & \multicolumn{2}{c}{Informer} \\
        \multicolumn{2}{c|}{} & \multicolumn{2}{c}{(Ours)} & \multicolumn{2}{c}{(2024)} & \multicolumn{2}{c}{(2023)} & \multicolumn{2}{c}{(2022)} & \multicolumn{2}{c}{(2021)} & \multicolumn{2}{c}{(2021)} \\
        \cmidrule(lr){1-2} \cmidrule(lr){3-4} \cmidrule(lr){5-6} \cmidrule(lr){7-8} \cmidrule(lr){9-10} \cmidrule(lr){11-12} \cmidrule(lr){13-14}
        \multicolumn{2}{c|}{Metric} & MSE & MAE & MSE & MAE & MSE & MAE & MSE & MAE & MSE & MAE & MSE & MAE \\
        \midrule
        \multirow{6}{*}{\rotatebox{90}{ETTh1}} & 96  & \underline{\textcolor{blue}{0.378}} & \textbf{\textcolor{red}{0.398}} & 0.386 & \underline{\textcolor{blue}{0.405}} & 0.423 & 0.448 & \textbf{\textcolor{red}{0.376}} & 0.419 & 0.449 & 0.459 & - & - \\
        & 192 & \underline{\textcolor{blue}{0.427}} & \textbf{\textcolor{red}{0.427}} & 0.441 & \underline{\textcolor{blue}{0.436}} & 0.471 & 0.474 & \textbf{\textcolor{red}{0.420}} & 0.448 & 0.500 & 0.482 & - & - \\
        & 336 & \underline{\textcolor{blue}{0.474}} & \textbf{\textcolor{red}{0.454}} & 0.487 & \underline{\textcolor{blue}{0.458}} & 0.570 & 0.546 & \textbf{\textcolor{red}{0.459}} & 0.465 & 0.521 & 0.496 & 1.128	& 0.873 \\
        & 720 & \textbf{\textcolor{red}{0.478}} & \textbf{\textcolor{red}{0.475}} & \underline{\textcolor{blue}{0.503}} & \underline{\textcolor{blue}{0.491}} & 0.653 & 0.621 & 0.506 & 0.507 & 0.514 & 0.512 & 1.215 & 0.896 \\
        \cmidrule(lr){2-14}
        & Avg & \textbf{\textcolor{red}{0.439}} & \textbf{\textcolor{red}{0.439}} & 0.454 & \underline{\textcolor{blue}{0.448}} & 0.529 & 0.522 & \underline{\textcolor{blue}{0.440}} & 0.460 & 0.496 & 0.487 & 1.172 & 0.885 \\
        \midrule
        \multirow{6}{*}{\rotatebox{90}{ETTh2}} & 96  & \textbf{\textcolor{red}{0.295}} & \textbf{\textcolor{red}{0.347}} & \underline{\textcolor{blue}{0.297}} & \underline{\textcolor{blue}{0.349}} & 0.745 & 0.584 & 0.358 & 0.397 & 0.346 & 0.388 & - & - \\
        & 192 & \textbf{\textcolor{red}{0.376}} & \textbf{\textcolor{red}{0.397}} & \underline{\textcolor{blue}{0.380}} & \underline{\textcolor{blue}{0.400}} & 0.877 & 0.656 & 0.429 & 0.439 & 0.456 & 0.452 & - & - \\
        & 336 & \textbf{\textcolor{red}{0.416}} & \textbf{\textcolor{red}{0.431}} & \underline{\textcolor{blue}{0.428}} & \underline{\textcolor{blue}{0.432}} & 1.043 & 0.731 & 0.496 & 0.487 & 0.482 & 0.486 & 2.723 & 1.340 \\
        & 720 & \textbf{\textcolor{red}{0.426}} & \textbf{\textcolor{red}{0.445}} & \underline{\textcolor{blue}{0.427}} & \underline{\textcolor{blue}{0.445}} & 1.104 & 0.763 & 0.463 & 0.474 & 0.515 & 0.511 & 3.467 & 1.473 \\
        \cmidrule(lr){2-14}
        & Avg & \textbf{\textcolor{red}{0.378}} & \textbf{\textcolor{red}{0.405}} & \underline{\textcolor{blue}{0.383}} & \underline{\textcolor{blue}{0.407}} & 0.942 & 0.684 & 0.437 & 0.449 & 0.450 & 0.459 & 3.095 & 1.407 \\
        \midrule
        \multirow{6}{*}{\rotatebox{90}{ETTm1}} & 96  & \textbf{\textcolor{red}{0.329}} & \underline{\textcolor{blue}{0.369}} & \underline{\textcolor{blue}{0.334}} & \textbf{\textcolor{red}{0.368}} & 0.404 & 0.426 & 0.379 & 0.419 & 0.505 & 0.475 & 0.678 & 0.614 \\
        & 192 & \textbf{\textcolor{red}{0.370}} & \textbf{\textcolor{red}{0.388}} & \underline{\textcolor{blue}{0.377}} & \underline{\textcolor{blue}{0.391}} & 0.450 & 0.451 & 0.426 & 0.441 & 0.553 & 0.496 & - & - \\
        & 336 & \textbf{\textcolor{red}{0.405}} & \textbf{\textcolor{red}{0.411}} & \underline{\textcolor{blue}{0.426}} & \underline{\textcolor{blue}{0.420}} & 0.532 & 0.515 & 0.445 & 0.459 & 0.621 & 0.537 & - & - \\
        & 720 & \textbf{\textcolor{red}{0.478}} & \textbf{\textcolor{red}{0.453}} & \underline{\textcolor{blue}{0.491}} & \underline{\textcolor{blue}{0.459}} & 0.666 & 0.589 & 0.543 & 0.490 & 0.671 & 0.561 & - & - \\
        \cmidrule(lr){2-14}
        & Avg & \textbf{\textcolor{red}{0.396}} & \textbf{\textcolor{red}{0.405}} & \underline{\textcolor{blue}{0.407}} & \underline{\textcolor{blue}{0.410}} & 0.513 & 0.496 & 0.448 & 0.452 & 0.588 & 0.517 & 0.678 & 0.614 \\
        \midrule
        \multirow{6}{*}{\rotatebox{90}{ETTm2}} & 96  & \textbf{\textcolor{red}{0.179}} & \textbf{\textcolor{red}{0.263}} & \underline{\textcolor{blue}{0.180}} & \underline{\textcolor{blue}{0.264}} & 0.287 & 0.366 & 0.203 & 0.287 & 0.255 & 0.339 & - & - \\
        & 192 & \textbf{\textcolor{red}{0.247}} & \textbf{\textcolor{red}{0.307}} & \underline{\textcolor{blue}{0.250}} & \underline{\textcolor{blue}{0.309}} & 0.414 & 0.492 & 0.269 & 0.328 & 0.281 & 0.340 & - & - \\
        & 336 & \textbf{\textcolor{red}{0.310}} & \textbf{\textcolor{red}{0.347}} & \underline{\textcolor{blue}{0.311}} & \underline{\textcolor{blue}{0.348}} & 0.597 & 0.542 & 0.325 & 0.366 & 0.339 & 0.372 & - & - \\
        & 720 & \textbf{\textcolor{red}{0.409}} & \textbf{\textcolor{red}{0.403}} & \underline{\textcolor{blue}{0.412}} & \underline{\textcolor{blue}{0.407}} & 1.730 & 1.042 & 0.421 & 0.415 & 0.433 & 0.432 & - & - \\
        \cmidrule(lr){2-14}
        & Avg & \textbf{\textcolor{red}{0.286}} & \textbf{\textcolor{red}{0.330}} & \underline{\textcolor{blue}{0.288}} & \underline{\textcolor{blue}{0.332}} & 0.757 & 0.610 & 0.305 & 0.349 & 0.327 & 0.371 & - & - \\
        \midrule
        \multirow{6}{*}{\rotatebox{90}{ECL}} & 96  & \textbf{\textcolor{red}{0.145}} & \textbf{\textcolor{red}{0.236}} & \underline{\textcolor{blue}{0.148}} & \underline{\textcolor{blue}{0.240}} & 0.219 & 0.314 & 0.193 & 0.308 & 0.201 & 0.317 & - & - \\
        & 192 & \textbf{\textcolor{red}{0.162}} & \textbf{\textcolor{red}{0.252}} & \underline{\textcolor{blue}{0.162}} & \underline{\textcolor{blue}{0.253}} & 0.231 & 0.322 & 0.201 & 0.315 & 0.222 & 0.334 & - & - \\
        & 336 & \textbf{\textcolor{red}{0.176}} & \textbf{\textcolor{red}{0.268}} & \underline{\textcolor{blue}{0.178}} & \underline{\textcolor{blue}{0.269}} & 0.246 & 0.337 & 0.214 & 0.329 & 0.231 & 0.338 & 0.381 & 0.431 \\
        & 720 & \textbf{\textcolor{red}{0.205}} & \textbf{\textcolor{red}{0.294}} & \underline{\textcolor{blue}{0.225}} & \underline{\textcolor{blue}{0.317}} & 0.280 & 0.363 & 0.246 & 0.355 & 0.254 & 0.361 & 0.406 & 0.443 \\
        \cmidrule(lr){2-14}
        & Avg & \textbf{\textcolor{red}{0.172}} & \textbf{\textcolor{red}{0.262}} & \underline{\textcolor{blue}{0.178}} & \underline{\textcolor{blue}{0.270}} & 0.244 & 0.334 & 0.214 & 0.327 & 0.227 & 0.338 & 0.394 & 0.437 \\
        \midrule
        \multirow{6}{*}{\rotatebox{90}{Exchange}} & 96  & \underline{\textcolor{blue}{0.087}} & \underline{\textcolor{blue}{0.206}} & \textbf{\textcolor{red}{0.086}} & \textbf{\textcolor{red}{0.206}} & 0.256 & 0.367 & 0.148 & 0.278 & 0.197 & 0.323 & - & - \\
        & 192 & \textbf{\textcolor{red}{0.176}} & \textbf{\textcolor{red}{0.299}} & \underline{\textcolor{blue}{0.177}} & \underline{\textcolor{blue}{0.299}} & 0.470 & 0.509 & 0.271 & 0.315 & 0.300 & 0.369 & - & - \\
        & 336 & \textbf{\textcolor{red}{0.327}} & \textbf{\textcolor{red}{0.415}} & \underline{\textcolor{blue}{0.331}} & \underline{\textcolor{blue}{0.417}} & 1.268 & 0.883 & 0.460 & 0.427 & 0.509 & 0.524 & - & - \\
        & 720 & \textbf{\textcolor{red}{0.833}} & \textbf{\textcolor{red}{0.689}} & \underline{\textcolor{blue}{0.847}} & \underline{\textcolor{blue}{0.691}} & 1.767 & 1.068 & 1.195 & 0.695 & 1.447 & 0.941 & - & - \\
        \cmidrule(lr){2-14}
        & Avg & \textbf{\textcolor{red}{0.356}} & \textbf{\textcolor{red}{0.402}} & \underline{\textcolor{blue}{0.360}} & \underline{\textcolor{blue}{0.403}} & 0.940 & 0.707 & 0.519 & 0.429 & 0.613 & 0.539 & - & - \\
        \midrule
        \multirow{6}{*}{\rotatebox{90}{Traffic}} & 96  & \textbf{\textcolor{red}{0.387}} & \textbf{\textcolor{red}{0.265}} & \underline{\textcolor{blue}{0.395}} & \underline{\textcolor{blue}{0.268}} & 0.522 & 0.290 & 0.587 & 0.366 & 0.613 & 0.388 & - & - \\
        & 192 & \textbf{\textcolor{red}{0.403}} & \textbf{\textcolor{red}{0.272}} & \underline{\textcolor{blue}{0.417}} & \underline{\textcolor{blue}{0.276}} & 0.530 & 0.293 & 0.604 & 0.373 & 0.616 & 0.382 & - & - \\
        & 336 & \textbf{\textcolor{red}{0.420}} & \textbf{\textcolor{red}{0.280}} & \underline{\textcolor{blue}{0.433}} & \underline{\textcolor{blue}{0.283}} & 0.558 & 0.305 & 0.621 & 0.383 & 0.622 & 0.337 & - & - \\
        & 720 & \textbf{\textcolor{red}{0.457}} & \textbf{\textcolor{red}{0.298}} & \underline{\textcolor{blue}{0.467}} & \underline{\textcolor{blue}{0.302}} & 0.589 & 0.328 & 0.626 & 0.382 & 0.660 & 0.408 & - & - \\
        \cmidrule(lr){2-14}
        & Avg & \textbf{\textcolor{red}{0.417}} & \textbf{\textcolor{red}{0.279}} & \underline{\textcolor{blue}{0.428}} & \underline{\textcolor{blue}{0.282}} & 0.550 & 0.304 & 0.610 & 0.376 & 0.628 & 0.379 & - & - \\
        \midrule
        \multirow{6}{*}{\rotatebox{90}{Weather}} & 96  & \underline{\textcolor{blue}{0.172}} & \textbf{\textcolor{red}{0.211}} & 0.174 & \underline{\textcolor{blue}{0.214}} & \textbf{\textcolor{red}{0.158}} & 0.230 & 0.217 & 0.296 & 0.266 & 0.336 & - & - \\
        & 192 & \underline{\textcolor{blue}{0.219}} & \textbf{\textcolor{red}{0.253}} & 0.221 & \underline{\textcolor{blue}{0.254}} & \textbf{\textcolor{red}{0.206}} & 0.277 & 0.276 & 0.336 & 0.307 & 0.367 & - & - \\
        & 336 & \underline{\textcolor{blue}{0.278}} & \underline{\textcolor{blue}{0.297}} & \underline{\textcolor{blue}{0.278}} & \textbf{\textcolor{red}{0.296}} & \textbf{\textcolor{red}{0.272}} & 0.335 & 0.339 & 0.380 & 0.359 & 0.395 & 0.702 & 0.620 \\
        & 720 & \textbf{\textcolor{red}{0.356}} & \textbf{\textcolor{red}{0.347}} & \underline{\textcolor{blue}{0.358}} & \underline{\textcolor{blue}{0.347}} & 0.398 & 0.418 & 0.403 & 0.428 & 0.419 & 0.428 & 0.831 & 0.731 \\
        \cmidrule(lr){2-14}
        & Avg & \textbf{\textcolor{red}{0.256}} & \textbf{\textcolor{red}{0.277}} & \underline{\textcolor{blue}{0.258}} & \underline{\textcolor{blue}{0.278}} & 0.259 & 0.315 & 0.309 & 0.360 & 0.338 & 0.382 & 0.767 & 0.676 \\
        \midrule
        \multirow{6}{*}{\rotatebox{90}{Solar-Energy}} & 96  & \textbf{\textcolor{red}{0.199}} & \textbf{\textcolor{red}{0.232}} & \underline{\textcolor{blue}{0.203}} & \underline{\textcolor{blue}{0.237}} & 0.310 & 0.331 & 0.242 & 0.342 & 0.884 & 0.711 & - & - \\
        & 192 & \textbf{\textcolor{red}{0.231}} & \underline{\textcolor{blue}{0.264}} & \underline{\textcolor{blue}{0.233}} & \textbf{\textcolor{red}{0.261}} & 0.734 & 0.725 & 0.285 & 0.380 & 0.834 & 0.692 & - & - \\
        & 336 & \textbf{\textcolor{red}{0.246}} & \textbf{\textcolor{red}{0.271}} & \underline{\textcolor{blue}{0.248}} & \underline{\textcolor{blue}{0.273}} & 0.750 & 0.735 & 0.282 & 0.376 & 0.941 & 0.723 & - & - \\
        & 720 & \underline{\textcolor{blue}{0.252}} & \underline{\textcolor{blue}{0.276}} & \textbf{\textcolor{red}{0.249}} & \textbf{\textcolor{red}{0.275}} & 0.769 & 0.765 & 0.357 & 0.427 & 0.882 & 0.717 & - & - \\
        \cmidrule(lr){2-14}
        & Avg & \textbf{\textcolor{red}{0.232}} & \textbf{\textcolor{red}{0.261}} & \underline{\textcolor{blue}{0.233}} & \underline{\textcolor{blue}{0.262}} & 0.641 & 0.639 & 0.291 & 0.381 & 0.885 & 0.711 & - & - \\
        \midrule
        \multicolumn{2}{c|}{$1^\text{st} \text{ Count}$} & \textbf{\textcolor{red}{37}} & \textbf{\textcolor{red}{40}} & \underline{\textcolor{blue}{2}} & \underline{\textcolor{blue}{5}} & 3 & 0 & 3 & 0 & 0 & 0 & 0 & 0 \\
        \bottomrule
    \end{tabular}
    }
\end{table*}

\begin{table*}[h!]
    \centering
    \caption{Performance comparison of different models on the PEMS datasets in terms of MSE and MAE across various prediction lengths \{12, 24, 48, 96\}.}

    \label{tab:pems-results}
    \scalebox{1}{ 
    \begin{tabular}{p{0.3cm} | c | c c | c c | c c | c c | c c}
        \toprule
        \multicolumn{2}{c|}{\multirow{2}{*}{Models}} &  \multicolumn{2}{c}{FISformer} & \multicolumn{2}{c}{ITransformer} & \multicolumn{2}{c}{Crossformer} & \multicolumn{2}{c}{Fedformer} & \multicolumn{2}{c}{Autoformer} \\
        \multicolumn{2}{c|}{} & \multicolumn{2}{c}{(Ours)} & \multicolumn{2}{c}{(2024)} & \multicolumn{2}{c}{(2023)} & \multicolumn{2}{c}{(2022)} & \multicolumn{2}{c}{(2021)} \\
        \cmidrule(lr){1-2} \cmidrule(lr){3-4} \cmidrule(lr){5-6} \cmidrule(lr){7-8} \cmidrule(lr){9-10} \cmidrule(lr){11-12}
        \multicolumn{2}{c|}{Metric} & MSE & MAE & MSE & MAE & MSE & MAE & MSE & MAE & MSE & MAE \\
        \midrule
        \multirow{7}{*}{\rotatebox{90}{PEMS03}}
        & 12 & \textbf{\textcolor{red}{0.068}} & \textbf{\textcolor{red}{0.174}} & \underline{\textcolor{blue}{0.071}} & \underline{\textcolor{blue}{0.174}} & 0.090 & 0.203 & 0.126 & 0.251 & 0.272 & 0.385 \\
        & 24 & \underline{\textcolor{blue}{0.095}} & \underline{\textcolor{blue}{0.204}} & \textbf{\textcolor{red}{0.093}} & \textbf{\textcolor{red}{0.201}} & 0.121 & 0.240 & 0.149 & 0.275 & 0.334 & 0.440 \\
        & 48 & \underline{\textcolor{blue}{0.153}} & \underline{\textcolor{blue}{0.262}} & \textbf{\textcolor{red}{0.125}} & \textbf{\textcolor{red}{0.236}} & 0.202 & 0.317 & 0.227 & 0.348 & 1.032 & 0.782 \\
        & 96 & \underline{\textcolor{blue}{0.235}} & \underline{\textcolor{blue}{0.336}} & \textbf{\textcolor{red}{0.164}} & \textbf{\textcolor{red}{0.275}} & 0.262 & 0.367 & 0.348 & 0.434 & 1.031 & 0.796 \\
        \cmidrule(lr){2-12}
        & Avg & \underline{\textcolor{blue}{0.138}} & \underline{\textcolor{blue}{0.244}} & \textbf{\textcolor{red}{0.113}} & \textbf{\textcolor{red}{0.222}} & 0.169 & 0.281 & 0.213 & 0.327 & 0.667 & 0.601 \\
        \midrule
        \multirow{7}{*}{\rotatebox{90}{PEMS04}}
        & 12 & \textbf{\textcolor{red}{0.076}} & \textbf{\textcolor{red}{0.180}} & \underline{\textcolor{blue}{0.078}} & \underline{\textcolor{blue}{0.183}} & 0.098 & 0.218 & 0.138 & 0.262 & 0.424 & 0.491 \\
        & 24 & \textbf{\textcolor{red}{0.093}} & \textbf{\textcolor{red}{0.202}} & \underline{\textcolor{blue}{0.095}} & \underline{\textcolor{blue}{0.205}} & 0.131 & 0.256 & 0.177 & 0.293 & 0.459 & 0.509 \\
        & 48 & \textbf{\textcolor{red}{0.118}} & \textbf{\textcolor{red}{0.230}} & \underline{\textcolor{blue}{0.120}} & \textbf{\textcolor{blue}{0.233}} & 0.205 & 0.326 & 0.270 & 0.368 & 0.646 & 0.610 \\
        & 96 & \textbf{\textcolor{red}{0.146}} & \textbf{\textcolor{red}{0.257}} & \underline{\textcolor{blue}{0.150}} & \underline{\textcolor{blue}{0.262}} & 0.402 & 0.457 & 0.341 & 0.427 & 0.912 & 0.748 \\
        \cmidrule(lr){2-12}
        & Avg & \textbf{\textcolor{red}{0.109}} & \textbf{\textcolor{red}{0.217}} & \underline{\textcolor{blue}{0.111}} & \underline{\textcolor{blue}{0.221}} & 0.209 & 0.314 & 0.231 & 0.337 & 0.610 & 0.590 \\
        \midrule
        \multirow{7}{*}{\rotatebox{90}{PEMS07}}
        & 12 & \textbf{\textcolor{red}{0.064}} & \textbf{\textcolor{red}{0.162}} & \underline{\textcolor{blue}{0.067}} & \underline{\textcolor{blue}{0.165}} & 0.094 & 0.200 & 0.109 & 0.225 & 0.199 & 0.336 \\
        & 24 & \textbf{\textcolor{red}{0.084}} & \textbf{\textcolor{red}{0.188}} & \underline{\textcolor{blue}{0.088}} & \underline{\textcolor{blue}{0.190}} & 0.139 & 0.247 & 0.125 & 0.244 & 0.323 & 0.420 \\
        & 48 & \textbf{\textcolor{red}{0.101}} & \textbf{\textcolor{red}{0.205}} & \underline{\textcolor{blue}{0.110}} & \underline{\textcolor{blue}{0.215}} & 0.311 & 0.369 & 0.165 & 0.288 & 0.390 & 0.470 \\
        & 96 & \textbf{\textcolor{red}{0.133}} & \textbf{\textcolor{red}{0.238}} & \underline{\textcolor{blue}{0.139}} & \underline{\textcolor{blue}{0.245}} & 0.396 & 0.442 & 0.262 & 0.376 & 0.554 & 0.578 \\
        \cmidrule(lr){2-12}
        & Avg & \textbf{\textcolor{red}{    0.095}} & \textbf{\textcolor{red}{0.198}} & \underline{\textcolor{blue}{0.101}} & \underline{\textcolor{blue}{0.204}} & 0.235 & 0.315 & 0.165 & 0.283 & 0.367 & 0.451 \\
        \midrule
        \multirow{7}{*}{\rotatebox{90}{PEMS08}}
        & 12 & \textbf{\textcolor{red}{0.078}} & \textbf{\textcolor{red}{0.179}} & \underline{\textcolor{blue}{0.079}} & \underline{\textcolor{blue}{0.182}} & 0.165 & 0.214 & 0.173 & 0.273 & 0.436 & 0.485 \\
        & 24 & \textbf{\textcolor{red}{0.110}} & \textbf{\textcolor{red}{0.211}} & \underline{\textcolor{blue}{0.115}} & \underline{\textcolor{blue}{0.219}} & 0.215 & 0.260 & 0.210 & 0.301 & 0.467 & 0.502 \\
        & 48 & \textbf{\textcolor{red}{0.179}} & \textbf{\textcolor{red}{0.228}} & \underline{\textcolor{blue}{0.186}} & \underline{\textcolor{blue}{0.235}} & 0.315 & 0.355 & 0.320 & 0.394 & 0.966 & 0.733 \\
        & 96 & \textbf{\textcolor{red}{0.213}} & \textbf{\textcolor{red}{0.254}} & \underline{\textcolor{blue}{0.221}} & \underline{\textcolor{blue}{0.267}} & 0.377 & 0.397 & 0.442 & 0.465 & 1.385 & 0.915 \\
        \cmidrule(lr){2-12}
        & Avg & \textbf{\textcolor{red}{0.145}} & \textbf{\textcolor{red}{0.218}} & \underline{\textcolor{blue}{0.150}} & \underline{\textcolor{blue}{0.226}} & 0.268 & 0.307 & 0.286 & 0.358 & 0.814 & 0.659 \\
        \midrule
        \multicolumn{2}{c|}{$1^\text{st} \text{ Count}$} & \textbf{\textcolor{red}{16}} & \textbf{\textcolor{red}{16}} & \underline{\textcolor{blue}{4}} & \underline{\textcolor{blue}{4}} & 0 & 0 & 0 & 0 & 0 & 0 \\
        \bottomrule
    \end{tabular}
    }
\end{table*}

In this section, we compare the forecasting performance of the proposed model with several state-of-the-art transformer-based methods for multivariate time series forecasting, including Informer~\cite{zhou2021informer}, Autoformer~\cite{wu2021autoformer}, FEDformer~\cite{zhou2022fedformer}, Crossformer~\cite{zhang2023crossformer}, iTransformer~\cite{liu2023itransformer}, and FANTF~\cite{chakraborty2025enhancing}. These models represent recent advances in efficient and effective sequence modeling for time series prediction, and serve as strong baselines for evaluating our approach. As our method incorporates the proposed FIS interaction mechanism within the transformer architecture, we refer to our model as FISformer throughout the remainder of this study.

\begin{figure*}[ht!]
\centering
\begin{tabular}{
>{\centering\arraybackslash}m{0.02\linewidth} 
@{\hspace{0.01cm}}
>{\centering\arraybackslash}m{0.32\linewidth}
 @{\hspace{0.05cm}}
>{\centering\arraybackslash}m{0.32\linewidth}
 @{\hspace{0.05cm}}
>{\centering\arraybackslash}m{0.32\linewidth}
}
     & Traffic & ECL & PEMS \\
    \rotatebox{90}{Self-Attention} &
    \includegraphics[width=0.95\linewidth]{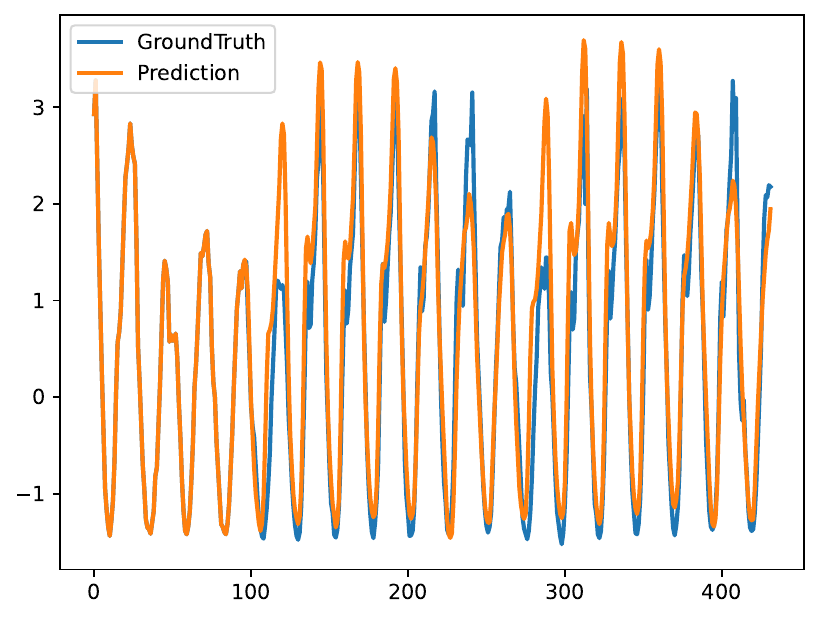} & 
    \includegraphics[width=0.95\linewidth]{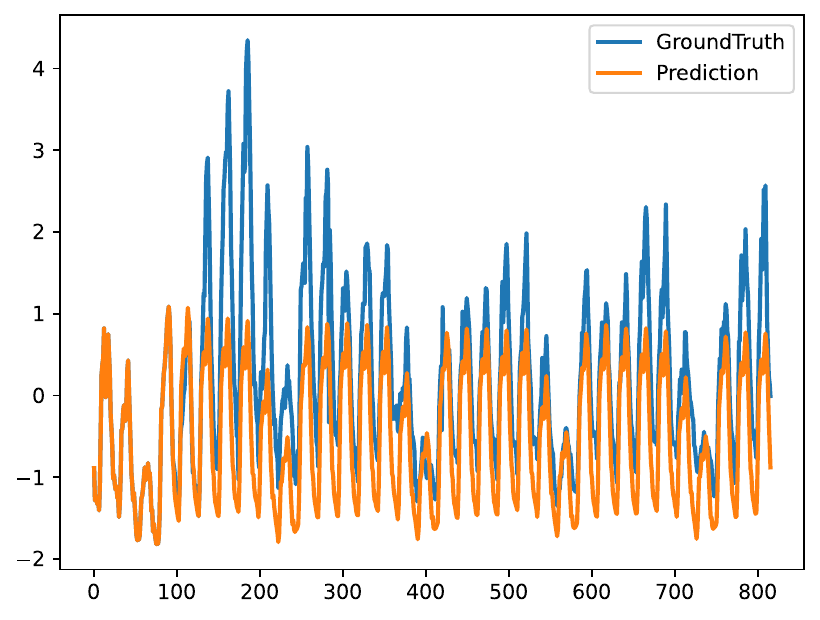} & 
    \includegraphics[width=0.95\linewidth]{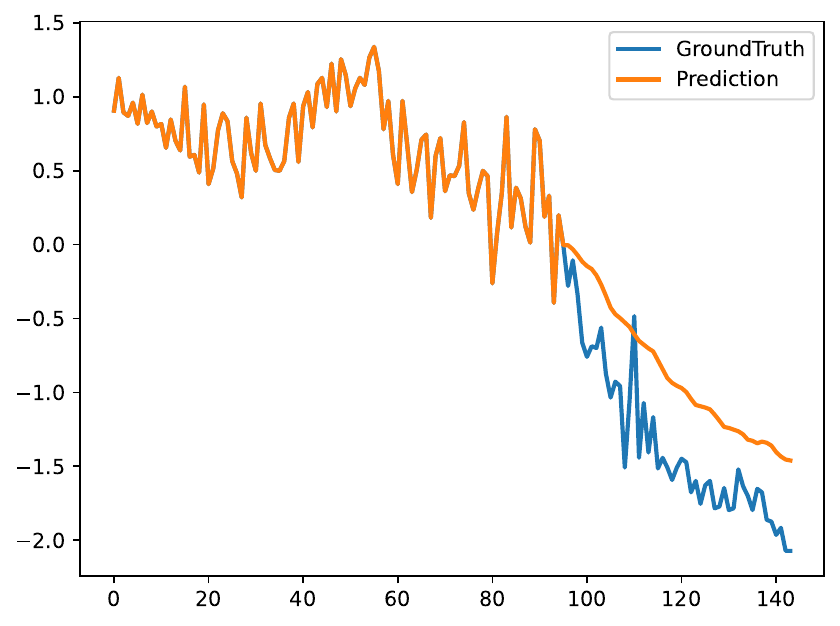} \\
    \rotatebox{90}{FIS Interaction} &
    \includegraphics[width=0.95\linewidth]{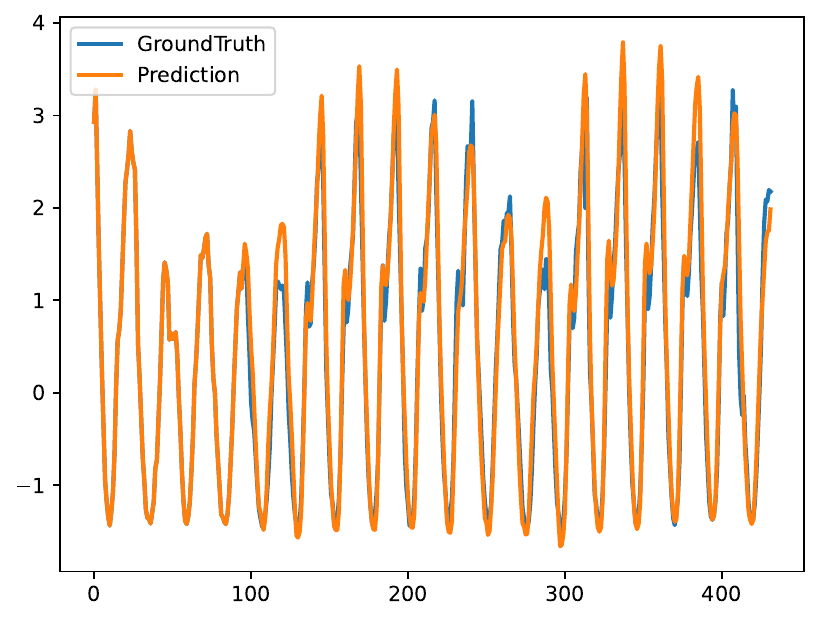} & 
    \includegraphics[width=0.95\linewidth]{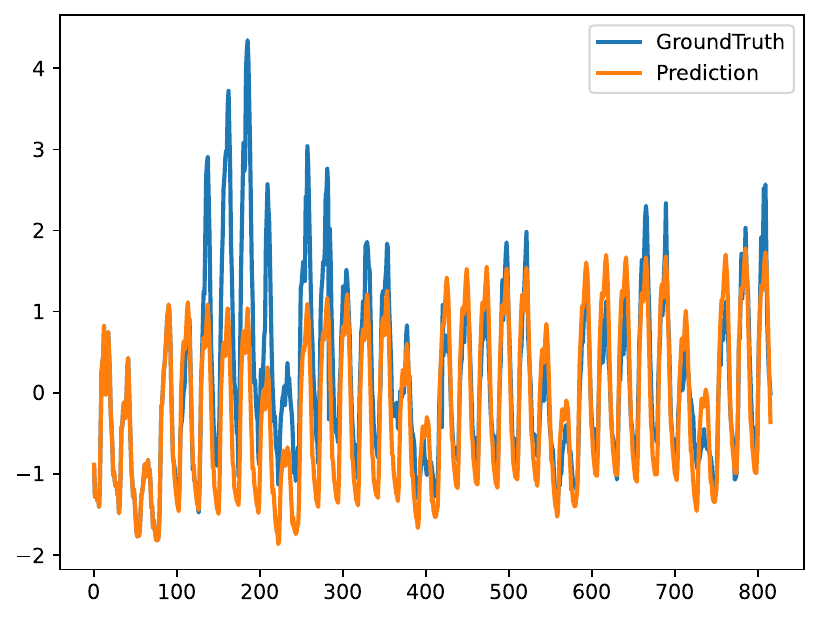} & 
    \includegraphics[width=0.95\linewidth]{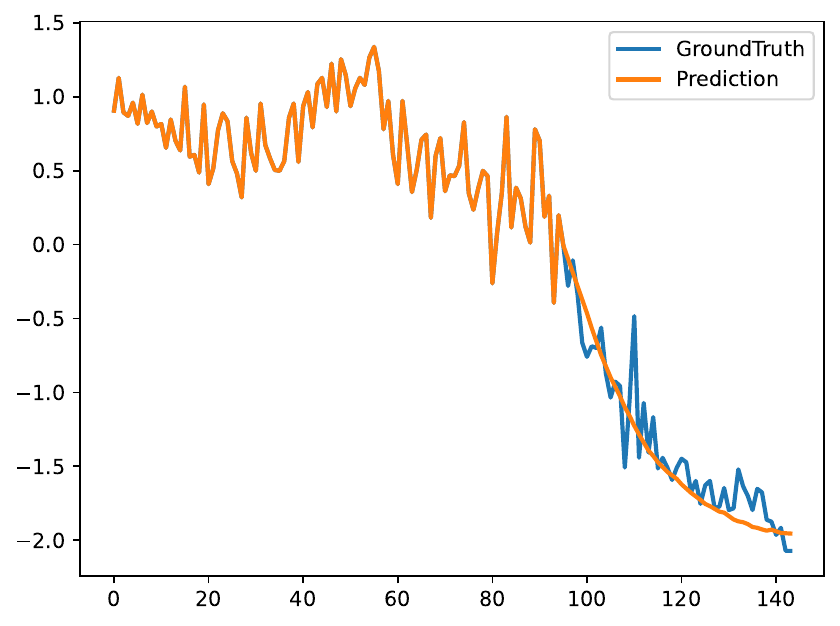} \\
\end{tabular}
\caption{Visualization of input-predict results on the Traffic, ECL and PEMS dataset.}
\label{fig:six_graphs}
\end{figure*}

The comparative forecasting results are presented in Table \ref{tab:results}. The best performance for each metric is highlighted in \textcolor{red}{red}, while the second-best scores are \underline{\textcolor{blue}{underlined in blue}} for clarity. We report both Mean Squared Error (MSE) and Mean Absolute Error (MAE), where lower values indicate better predictive performance. As observed in the table, the proposed model, FISformer, achieves the best performance on 6 out of 7 datasets for both MSE and MAE. Furthermore, it ranks second on the remaining 1 dataset across each metric. ITransformer ranks as the second-best method after our proposed approach. To offer a comprehensive comparison between the proposed \textit{FISformer} and the baseline \textit{iTransformer}, we present additional visualized forecasting results for three representative datasets, as illustrated in Figure~\ref{fig:six_graphs}. These examples visually highlight the predictive performance and behavior of the proposed model across various time series patterns. As clearly seen in the PEMS sample, our model effectively forecasts by closely following the central trend of the time series, demonstrating its ability to capture both short-term fluctuations and overall temporal structure.

To provide a more detailed analysis, we elaborate on the forecasting results for each dataset and prediction length in Tables~\ref{tab:all-results} and~\ref{tab:pems-results}. Table~\ref{tab:pems-results} presents the forecasting performance on the four public subsets from the PEMS dataset across various sequence lengths. Table~\ref{tab:all-results} provides the detailed results for all prediction lengths across the nine widely acknowledged multivariate time series forecasting benchmark datasets.

It is observed that the competitive baseline, iTransformer, generally yields worse forecasting performance compared to our proposed model, particularly on low-dimensional time series such as those in the ETT dataset. In contrast, FISformer consistently demonstrates improved accuracy across both high- and low-dimensional hourly time series. Other transformer-based methods tend to underperform relative to both iTransformer and FISformer. This performance discrepancy can be attributed to their use of time-independent tokenization strategies, which may neglect crucial temporal dependencies. In contrast, the inherent design of the Transformer architecture is well-suited for capturing temporal dynamics and modeling inter-variable correlations. 

It is confirmed that the inverted architecture introduced by iTransformer effectively leverages the strengths of Transformer-based models in capturing temporal dependencies and multivariate correlations. Our proposed FIS Interaction mechanism integrates seamlessly with this architecture, further enhancing forecasting performance. Moreover, it is noteworthy that the proposed FIS interaction mechanism is also compatible with time-dependent tokenization strategies, as demonstrated by the results presented in the corresponding Table~\ref{tab:informer}. This flexibility highlights the adaptability and robustness of the FIS interaction mechanism across different tokenization schemes.

To further evaluate the effectiveness of the proposed \textit{FIS Interaction} mechanism, we compare our model against the Fuzzy Attention-Integrated Transformer (FANTF)~\cite{chakraborty2025enhancing}, which introduces fuzzification into the attention computation process. 
While FANTF employs a fuzzy membership mapping over the input token embeddings to enhance robustness, its approach is limited to the fuzzification stage of the attention mechanism. 
In contrast, the proposed FISformer integrates a complete first-order Sugeno Fuzzy Inference System (FIS) into the interaction computation, encompassing fuzzification, rule generation, and defuzzification. 
This allows FISformer to model nonlinear and uncertainty-aware relationships among tokens through rule-based reasoning rather than heuristic fuzzification alone. 
As shown in Table~\ref{tab:comparison_fantf}, FISformer consistently outperforms FANTF across all benchmark datasets, demonstrating superior adaptability under noisy temporal conditions and improved generalization to multivariate dependencies. 
These results confirm that embedding a full fuzzy inference pipeline provides a more expressive and interpretable formulation than partial fuzzification of token representations.

\begin{table}[h]
    \centering
    \caption{Quantitative comparison between FANTF~\cite{chakraborty2025enhancing} and FISformer on benchmark datasets in terms of MSE and MAE at a prediction length of 192. “Promotion” denotes the relative improvement of FISFormer over FANTF.}

    \label{tab:comparison_fantf}
    \scalebox{1}{ 
    \begin{tabular}{c | c c | c c | c c}
        \toprule
        \multirow{3}{*}{Models} &  \multicolumn{2}{c}{FISformer} & \multicolumn{2}{c}{FANTF} & \multicolumn{2}{c}{\multirow{2}{*}{Promotion (\%)}} \\
         & \multicolumn{2}{c}{(Ours)} & \multicolumn{2}{c}{(2025)} & \multicolumn{2}{c}{} \\
        \cmidrule(lr){1-1} \cmidrule(lr){2-3} \cmidrule(lr){4-5} \cmidrule(lr){6-7}
        Metric & MSE & MAE & MSE & MAE & MSE & MAE \\
        \midrule
        \makecell{ETTh1}  & \textbf{0.427} & \textbf{0.427} & 0.456 & 0.446 & \textcolor{mygreen}{6.8\%} & \textcolor{mygreen}{4.4\%} \\
        \midrule
        \makecell{ETTh2} & \textbf{0.375} & \textbf{0.397} & 0.387 & 0.408 & \textcolor{mygreen}{3.2\%} & \textcolor{mygreen}{2.8\%} \\
        \midrule
        \makecell{ETTm1}  & \textbf{0.370} & \textbf{0.388} & 0.409 & 0.409 & \textcolor{mygreen}{10.5\%} & \textcolor{mygreen}{5.4\%}\\
        \midrule
        \makecell{ETTm2}   & \textbf{0.247} & \textbf{0.307} & 0.291 & 0.334 & \textcolor{mygreen}{17.8\%} & \textcolor{mygreen}{8.8\%} \\
        \midrule
        \makecell{ECL} & \textbf{0.162} & \textbf{0.252} & 0.183 & 0.272 & \textcolor{mygreen}{13.0\%} & \textcolor{mygreen}{7.9\%} \\
        \midrule
        \makecell{Exchange} & \textbf{0.176} & \textbf{0.299} & 0.361 & 0.404 & \textcolor{mygreen}{105.1\%} & \textcolor{mygreen}{35.1\%} \\
        \midrule
        \makecell{Traffic} & \textbf{0.403} & \textbf{0.272} & 0.440 & 0.294 & \textcolor{mygreen}{9.2\%} & \textcolor{mygreen}{8.1\%} \\
        \midrule
        \makecell{Weather}   & \textbf{0.217} & \textbf{0.252} & 0.251 & 0.279 & \textcolor{mygreen}{15.7\%} & \textcolor{mygreen}{10.7\%} \\
        \bottomrule
    \end{tabular}
    }
\end{table}

\subsection{Ablation Study}

\textit{Effect of FIS Interaction.} To evaluate the effectiveness of the proposed FIS interaction mechanism, we replaced the standard attention with FIS interaction module and present the results in Table~\ref{tab:fullfuzzy}. As shown, the proposed mechanism yields performance improvements in most cases. Furthermore, we integrated FIS interaction into the Informer architecture \cite{zhou2021informer}, which utilizes time-wise tokenization by replacing its original ProbSparse Attention. The corresponding results, reported in Table~\ref{tab:informer}, demonstrate that FIS interaction generalizes well across architectures and consistently enhances forecasting accuracy.

\begin{table*}[h]
    \centering
    \caption{Effect of FIS Interaction compared to standard Self-Attention in terms of MSE and MAE across different datasets.}

    \label{tab:fullfuzzy}
    \scalebox{1}{
    \setlength{\tabcolsep}{4pt}
    \begin{tabular}{c | c c | c c | c c | c c | c c | c c | c c}
        \toprule
        Datasets &  \multicolumn{2}{c}{ECL} & \multicolumn{2}{c}{ETT (Avg)} & \multicolumn{2}{c}{Exchange} & \multicolumn{2}{c}{Traffic} & \multicolumn{2}{c}{Weather} & \multicolumn{2}{c}{Solar-Energy} & \multicolumn{2}{c}{PEMS (Avg)} \\
        \cmidrule(lr){2-3} \cmidrule(lr){4-5} \cmidrule(lr){6-7} \cmidrule(lr){8-9} \cmidrule(lr){10-11} \cmidrule(lr){12-13} \cmidrule(lr){14-15}
        Metric & MSE & MAE & MSE & MAE & MSE & MAE & MSE & MAE & MSE & MAE & MSE & MAE & MSE & MAE \\
        \midrule
        \makecell{w/Self-Attention}    
        & 0.178 & 0.270 & 0.383 & 0.399 & 0.360 & 0.403 & 0.428 & 0.282 & 0.258 & 0.278 & 0.233 & 0.262 & \textbf{0.119} & \textbf{0.218} \\
        \makecell{\textbf{w/FIS Interaction}} 
        & \textbf{0.172} & \textbf{0.262} & \textbf{0.375} & \textbf{0.395} & \textbf{0.356} & \textbf{0.402} & \textbf{0.417} & \textbf{0.279} & \textbf{0.256} & \textbf{0.277} & \textbf{0.232} & \textbf{0.261} & 0.122 & 0.219 \\
        \midrule
        \makecell{Promotion}
        & \textcolor{mygreen}{3.7\%} & \textcolor{mygreen}{2.8\%} & \textcolor{mygreen}{2.2\%} & \textcolor{mygreen}{1.1\%} & \textcolor{mygreen}{1.2\%} & \textcolor{mygreen}{0.2\%} & \textcolor{mygreen}{2.7\%} & \textcolor{mygreen}{1.3\%} & \textcolor{mygreen}{0.9\%} & \textcolor{mygreen}{0.2\%} & \textcolor{mygreen}{0.6\%} & \textcolor{mygreen}{0.3\%} & \textcolor{red}{-2.3\%} & \textcolor{red}{-0.7\%} \\
        \bottomrule
    \end{tabular}
    }
\end{table*}

\begin{table}[ht!]
    \centering
    \caption{The effect of integrating FIS Interaction into the Informer model by replacing the original ProbSparse Attention mechanism. Results are reported in terms of MSE and MAE across various datasets.}

    \label{tab:informer}
    \scalebox{1}{
    \setlength{\tabcolsep}{4pt}
    \begin{tabular}{p{0.4cm} | c | c c | c c | c c}
        \toprule
        \multicolumn{2}{c|}{\multirow{3}{*}{Models}} &  \multicolumn{2}{c|}{Informer} & \multicolumn{2}{c|}{Informer} & \multicolumn{2}{c}{\multirow{3}{*}{Promotion (\%)}} \\
        \multicolumn{2}{c|}{} & \multicolumn{2}{c|}{ProbAttention} & \multicolumn{2}{c|}{FIS Interaction} & \multicolumn{2}{c}{} \\
        \multicolumn{2}{c|}{} & \multicolumn{2}{c|}{(2021)} & \multicolumn{2}{c|}{(Ours)} & \multicolumn{2}{c}{} \\
        \cmidrule(lr){1-2} \cmidrule(lr){3-4} \cmidrule(lr){5-6} \cmidrule(lr){7-8}
        \multicolumn{2}{c|}{Metric} & MSE & MAE & MSE & MAE & MSE & MAE \\
        \midrule
        \multirow{6}{*}{\rotatebox{90}{ETTh1}} 
        & 24 & 0.577 & 0.549 & 0.435 & 0.472 & \textcolor{mygreen}{24.6\%} & \textcolor{mygreen}{14.1\%} \\
        & 48 & 0.685 & 0.625 & 0.623 & 0.593 & \textcolor{mygreen}{9.1\%} & \textcolor{mygreen}{5.1\%} \\
        & 168 & 0.931 & 0.752 & 0.817 & 0.714 & \textcolor{mygreen}{12.3\%} & \textcolor{mygreen}{5.1\%} \\
        & 336 & 1.128 & 0.873 & 0.955 & 0.781 & \textcolor{mygreen}{15.4\%} & \textcolor{mygreen}{10.5\%} \\
        & 720 & 1.215 & 0.896 & 1.150 & 0.889 & \textcolor{mygreen}{5.4\%} & \textcolor{mygreen}{0.8\%} \\
        \cmidrule(lr){2-8}
        & AVG & 0.907 & 0.739 & 0.796 & 0.690 & \textcolor{mygreen}{12.3\%} & \textcolor{mygreen}{6.7\%} \\
        \midrule
        \multirow{6}{*}{\rotatebox{90}{ETTh2}} 
        & 24 & 0.720 & 0.665 & 0.423 & 0.505 & \textcolor{mygreen}{41.3\%} & \textcolor{mygreen}{24.1\%} \\
        & 48 & 1.457 & 1.001 & 1.389 & 0.972 & \textcolor{mygreen}{4.7\%} & \textcolor{mygreen}{2.9\%} \\
        & 168 & 3.489 & 1.515 & 2.863 & 1.361 & \textcolor{mygreen}{18.0\%} & \textcolor{mygreen}{10.2\%} \\
        & 336 & 2.723 & 1.340 & 2.432 & 1.285 & \textcolor{mygreen}{10.7\%} & \textcolor{mygreen}{4.1\%} \\
        & 720 & 3.467 & 1.473 & 2.912 & 1.357 & \textcolor{mygreen}{16.0\%} & \textcolor{mygreen}{7.8\%} \\
        \cmidrule(lr){2-8}
        & AVG & 2.371 & 1.199 & 2.004 & 1.096 & \textcolor{mygreen}{15.5\%} & \textcolor{mygreen}{8.6\%} \\
        \midrule
        \multirow{6}{*}{\rotatebox{90}{ETTm1}} 
        & 24 & 0.323 & 0.369 & 0.293 & 0.368 & \textcolor{mygreen}{9.2\%} & \textcolor{mygreen}{0.3\%} \\
        & 48 & 0.494 & 0.503 & 0.476 & 0.453 & \textcolor{mygreen}{3.6\%} & \textcolor{mygreen}{9.9\%} \\
        & 96 & 0.678 & 0.614 & 0.481 & 0.478 & \textcolor{mygreen}{29.1\%} & \textcolor{mygreen}{22.2\%} \\
        & 288 & 1.056 & 0.786 & 0.748 & 0.659 & \textcolor{mygreen}{29.2\%} & \textcolor{mygreen}{16.1\%} \\
        & 672 & 1.192 & 0.926 & 0.961 & 0.784 & \textcolor{mygreen}{19.4\%} & \textcolor{mygreen}{15.4\%} \\
        \cmidrule(lr){2-8}
        & AVG & 0.749 & 0.640 & 0.592 & 0.548 & \textcolor{mygreen}{20.9\%} & \textcolor{mygreen}{14.3\%} \\
        \midrule
        \multirow{6}{*}{\rotatebox{90}{ECL}} 
        & 48 & 0.344 & 0.393 & 0.278 & 0.368 & \textcolor{mygreen}{7.6\%} & \textcolor{mygreen}{5.1\%} \\
        & 168 & 0.368 & 0.424 & 0.280 & 0.376 & \textcolor{mygreen}{7.5\%} & \textcolor{mygreen}{10.4\%} \\
        & 336 & 0.381 & 0.431 & 0.286 & 0.379 & \textcolor{mygreen}{11.7\%} & \textcolor{mygreen}{5.9\%} \\
        & 720 & 0.406 & 0.443 & 0.281 & 0.373 & \textcolor{mygreen}{13.1\%} & \textcolor{mygreen}{6.3\%} \\
        & 960 & 0.460 & 0.548 & 0.293 & 0.373 & \textcolor{mygreen}{29.2\%} & \textcolor{mygreen}{23.2\%} \\
        \cmidrule(lr){2-8}
        & AVG & 0.392 & 0.448 & 0.284 & 0.374 & \textcolor{mygreen}{16.0\%} & \textcolor{mygreen}{11.2\%} \\
        \midrule
        \multirow{6}{*}{\rotatebox{90}{Weather}} 
        & 24 & 0.335 & 0.381 & 0.309 & 0.362 & \textcolor{mygreen}{19.1\%} & \textcolor{mygreen}{6.3\%} \\
        & 48 & 0.395 & 0.459 & 0.365 & 0.411 & \textcolor{mygreen}{24.0\%} & \textcolor{mygreen}{11.3\%} \\
        & 168 & 0.608 & 0.567 & 0.537 & 0.533 & \textcolor{mygreen}{25.0\%} & \textcolor{mygreen}{12.1\%} \\
        & 336 & 0.702 & 0.620 & 0.610 & 0.581 & \textcolor{mygreen}{30.7\%} & \textcolor{mygreen}{15.8\%} \\
        & 720 & 0.831 & 0.731 & 0.589 & 0.562 & \textcolor{mygreen}{36.3\%} & \textcolor{mygreen}{31.9\%} \\
        \cmidrule(lr){2-8}
        & AVG & 0.574 & 0.552 & 0.482 & 0.490 & \textcolor{mygreen}{27.6\%} & \textcolor{mygreen}{16.5\%} \\
        \midrule
        \multicolumn{6}{r|}{\textbf{Overall Promotion (\%)}} & \textbf{\textcolor{mygreen}{18.1\%}} & \textbf{\textcolor{mygreen}{11.1\%}}  \\      
        \bottomrule
    \end{tabular}
    }
\end{table}

\subsection{Impact of Learning Rate}

\begin{figure}[ht!]
    \centering
\includegraphics[width=1.0\linewidth]{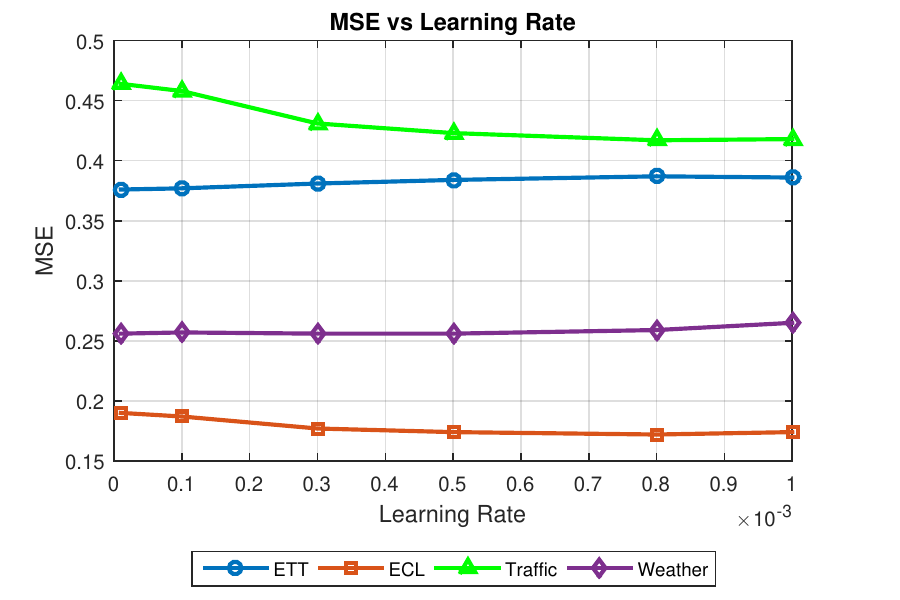}
    \caption{Impact of learning rate on MSE scores for the ETT (Avg), ECL, Traffic, and Weather datasets. }
    \label{fig:lr}
\end{figure}

The learning rate is a critical hyperparameter that significantly influences deep learning models' convergence and generalization ability. To analyze its impact to evaluate learning rate sensitivity, we conduct a series of experiments by varying the learning rate while keeping all other settings fixed for different datasets. The results given in Figure~\ref{fig:lr} reveal that the model is generally not highly sensitive to small changes in the learning rate. For some datasets, such as Weather and ETT (Avg), smaller learning rates yield slightly better performance, while for others, including Traffic and ECL, relatively larger learning rates are more effective. These results suggest that while the model is robust to moderate fluctuations in the learning rate, dataset-specific tuning can further enhance performance.

\subsection{Analysis of Different Membership Functions (MF)}

\begin{table}[h]
    \centering
    \caption{Analysis of different Membership Functions (MFs)—Gaussian, Triangular, and Trapezoidal—across various datasets.}

    \label{tab:mftype}
    \scalebox{1}{ 
    \begin{tabular}{c | c c | c c | c c}
        \toprule
        \multirow{2}{*}{MFs Types /} &  \multicolumn{2}{c}{Gaussian} & \multicolumn{2}{c}{Triangular} & \multicolumn{2}{c}{Trapezoidal}\\
        \cmidrule(lr){2-3} \cmidrule(lr){4-5} \cmidrule(lr){6-7}
        Metric & MSE & MAE & MSE & MAE & MSE & MAE \\
        \midrule
        \makecell{ECL}  & \textbf{0.172} & \textbf{0.262} & 0.176 & 0.267 & 0.175 & 0.267 \\
        \midrule
        \makecell{ETT (Avg)} & \textbf{0.375} & \textbf{0.395} & 0.383 & 0.400 & 0.385 & 0.400 \\
        \midrule
        \makecell{Exchange}  & \textbf{0.356} & \textbf{0.402} & 0.356 & 0.406 & 0.362 & 0.408 \\
        \midrule
        \makecell{Traffic}   & \textbf{0.417} & \textbf{0.279} & 0.422 & 0.281 & 0.423 & 0.282 \\
        \midrule
        \makecell{Weather}   & \textbf{0.256} & \textbf{0.277} & 0.258 & 0.279 & 0.259 & 0.279 \\
        \midrule
        \makecell{Solar-Energy} & \textbf{0.232} & \textbf{0.261} & 0.237 & 0.261 & 0.239 & 0.261 \\
        \midrule
        \makecell{PEMS (Avg)} & \textbf{0.122} & \textbf{0.219} & 0.136 & 0.233 & 0.136 & 0.232 \\
        \bottomrule
    \end{tabular}
    }
\end{table}

Membership functions (MFs) are a key component of the Fuzzy Inference System, as they define how input values are mapped to degrees of membership. To investigate their effect, we evaluate the model using different types of MFs, including triangular, trapezoidal, and Gaussian functions. The results, given in Table~\ref{tab:mftype} indicate that Gaussian membership functions generally yield superior performance due to their smoothness and better capability to represent gradual transitions in temporal data. However, triangular and trapezoidal MFs offer comparable performance with reduced computational complexity, suggesting a trade-off between accuracy and efficiency.

\subsection{Time Complexity}
We analyze the computational complexity of the proposed \textit{FIS Interaction} mechanism and compare it against standard self-attention to assess scalability and efficiency.

Table~\ref{tab:complexity-comparison} summarizes the asymptotic complexity of both methods. In contrast to self-attention, which requires pairwise token interactions leading to a quadratic cost of \( \mathcal{O}(T^2 \cdot d) \), the proposed FIS Interaction performs fuzzy reasoning independently along the token axis, resulting in a linear complexity with respect to sequence length. Specifically, fuzzification and rule-based defuzzification are applied per token–feature pair, producing a total cost of \( \mathcal{O}(T \cdot d \cdot R) \), where \(R\) denotes the number of fuzzy rules (or membership functions). The remaining projection operations preserve the standard linear complexity term \( \mathcal{O}(T \cdot d^2) \).

\begin{table}[!h!]
\centering
\caption{Comparison of time complexity for Self-Attention and FIS Interaction. \( T \) is sequence length, \( d \) is feature dimension, and \( R \) is the number of membership functions.}
\label{tab:complexity-comparison}
\begin{tabular}{lcc}
\toprule
\textbf{Component} & \textbf{Self-Attention} & \textbf{FIS Interaction} \\
\midrule
Q/K/V projection 
& \( \mathcal{O}(T \cdot d^{2}) \) 
& \( \mathcal{O}(T \cdot d^{2}) \) \\

Self-Attention
& \( \mathcal{O}(T^{2} \cdot d) \) 
& -- \\

Fuzzification
& -- 
& \( \mathcal{O}(T \cdot d \cdot R) \) \\

Rule generation \& \\ Defuzzification
& -- 
& \( \mathcal{O}(T \cdot d \cdot R) \) \\

\midrule
\textbf{Time Complexity}  
& \( \mathcal{O}(T^{2} \cdot d) \) 
& \( \mathcal{O}(T \cdot d \cdot R) \) \\
\bottomrule
\end{tabular}
\end{table}

As shown in Table~\ref{tab:complexity-comparison}, the proposed \textit{FIS Interaction} 
maintains a comparable computational complexity to standard self-attention while introducing 
significant advantages in interpretability and modeling flexibility. 
By embedding fuzzy inference into the interaction process, the mechanism captures both 
localized and holistic dependencies through rule-based reasoning. 
Notably, when the number of fuzzy rules \( R \ll T \), i.e., much smaller than the sequence length, 
the \textit{FIS Interaction} becomes substantially more efficient than quadratic attention mechanisms. 
This demonstrates that the proposed approach serves as a scalable and efficient alternative 
to conventional attention, particularly in domains where interpretability and uncertainty-aware 
representations are crucial.

\section{Conclusion}

In this work, we introduced \textit{FISformer}, a novel transformer architecture that integrates a 
Fuzzy Inference System (FIS) into the attention computation process for multivariate time series forecasting. 
Unlike traditional self-attention mechanisms that rely on deterministic dot-product similarity, 
the proposed \textit{FIS Interaction} replaces this operation with a rule-based fuzzy reasoning procedure, 
allowing the model to capture nonlinear, uncertain, and context-dependent relationships among tokens. 
By employing learnable membership functions and first-order Sugeno fuzzy rules, FISformer effectively 
models uncertainty and enhances interpretability while preserving the expressive capacity of transformer-based architectures.

Comprehensive experiments on multiple benchmark datasets demonstrate that 
FISformer achieves competitive or superior forecasting performance compared to conventional transformer models, 
while maintaining comparable computational complexity. 
These results validate the effectiveness of integrating fuzzy inference mechanisms into the 
token interaction process, leading to more robust, adaptable, and interpretable forecasting models.

This study highlights the potential of combining fuzzy logic with deep neural sequence models 
as a general framework for uncertainty-aware representation learning. 
Future research directions include extending FISformer to cross-modal and graph-based forecasting tasks, 
as well as exploring hybrid neuro-fuzzy architectures for enhanced explainability and domain adaptability 
in broader sequential and structured data applications.

\bibliographystyle{ieeetr}  
\bibliography{references} 

\end{document}